\documentclass[a4paper, 10pt, conference]{ieeeconf}      % Use this line for a4
\usepackage{FG2018}

\IEEEoverridecommandlockouts                              % This command is only
\usepackage{times}
\usepackage{epsfig}
\usepackage{graphicx}
\usepackage{amsmath}
\usepackage{amssymb}
\usepackage{threeparttable}
\usepackage{multirow}
\usepackage{comment}
\usepackage{subfigure}
\usepackage{color}
\usepackage{booktabs}

\usepackage{hyperref}
\hypersetup{
	breaklinks=true,
	bookmarks=false,
	colorlinks=true,
	linkcolor=red,
}

\def\FGPaperID{21} % *** Enter the FG 2018 Paper ID here

\title{\LARGE \bf
	PersonRank: Detecting Important People in Images
}

\author{\parbox{16cm}{\centering
		{\large Wei-Hong Li$^\dagger$, Benchao Li$^\dagger$ and Wei-Shi Zheng$^{\ddagger *}$ \thanks{*Corresponding author.}}\\
		{\normalsize
			$^\dagger$ School of Electronics and Information Technology, Sun Yat-Sen University, GuangZhou, China\\
			$^\ddagger$ School of Data and Computer Science, Sun Yat-Sen University, GuangZhou, China} }\\
}

\newcommand{\weihong}{\textcolor[rgb]{0.00,0.00,0.00}}
\newcommand{\weihongnew}{\textcolor[rgb]{0.00,0.00,0.00}}
\newcommand{\whnew}{\textcolor[rgb]{0.00,0.00,0.00}}

\newcommand{\WHnew}{\textcolor[rgb]{0.00,0.00,0.00}}

\newcommand{\whll}{\textcolor[rgb]{0.00,0.00,0.00}}
\newcommand{\whlll}{\textcolor[rgb]{0.00,0.00,0.00}}
\newcommand{\wwhl}{\textcolor[rgb]{0.00,0.00,0.00}}
\newcommand{\whllll}{\textcolor[rgb]{0.00,0.00,0.00}}
\newcommand{\whlllll}{\textcolor[rgb]{0.00,0.00,0.00}}
\newcommand{\seccolor}{\textcolor[rgb]{1.00,0.30,0.00}}
\newcommand{\wss}{\textcolor[rgb]{0.00,0.00,0.00}}
\newcommand{\wsss}{\textcolor[rgb]{0.00,0.00,0.00}}
\newcommand{\WH}{\textcolor[rgb]{0.00,0.00,0.00}}

\renewcommand{\tabcolsep}{0.025cm}

\begin{document}
	\FGfinalcopy
	\ifFGfinal
	\thispagestyle{empty}
	\pagestyle{empty}
	\else
	\author{Anonymous FG 2018 submission\\ Paper ID \FGPaperID \\}
	\pagestyle{plain}
	\fi
	\maketitle
	
	%%%%%%%%% TITLE
%	\title{PersonRank: Detecting Important People in Images}
%	
%	\author{First Author\\
%		Institution1\\
%		Institution1 address\\
%		{\tt\small firstauthor@i1.org}
%		% For a paper whose authors are all at the same institution,
%		% omit the following lines up until the closing ``}''.
%		% Additional authors and addresses can be added with ``\and'',
%		% just like the second author.
%		% To save space, use either the email address or home page, not both
%		\and Second Author\\ Institution2\\ First line of institution2 address\\
%		{\tt\small secondauthor@i2.org} }
%	
%	\maketitle
%	\thispagestyle{empty}
	
	%%%%%%%%% ABSTRACT
	\begin{abstract}
		
		\whlll{
			Always, some individuals in images are more important/attractive than others in some events such as presentation, basketball game or speech. 
			\WH{However, it is challenging to find important people among all individuals in images directly based on their spatial or appearance information due to the existence of diverse variations of pose, action,  appearance of persons and various changes of occasions.
%				Another challenge remained unsolved is to infer the most important people from multiple cues, including visual and spatial information.
				%		However, it is not intuitive for a machine to find out them among large amount of visual cues against diverse changes of pose, action, and appearance of persons. 
				We overcome this difficulty by constructing a multiple Hyper-Interaction Graph to treat each individual in an image as a node and inferring the most active node referring to interactions estimated by various types of clews.
				%		casting the whole problem as finding the most active node.
				%For each of four types of clues, 
				We model pairwise interactions between persons as the edge message communicated between nodes, resulting in a bidirectional pairwise-interaction graph. 
				To enrich the person-person interaction estimation, we further introduce a unidirectional hyper-interaction graph that models the consensus of interaction between a focal person and any person in a local region around.
				Finally, we modify the PageRank algorithm to infer the activeness of persons on the multiple Hybrid-Interaction Graph (HIG), the union of the pairwise-interaction and hyper-interaction graphs, and we call our algorithm the \emph{PersonRank}.}
			%		To overcome this difficulty, it is the first time to treat each individual in an image as a node in a graph and cast the whole problem as finding the most active node. 
			%		We model four types of \wsss{pairwise} interactions between persons as the edge message \wwhl{communicated} between nodes, and this results in a \wwhl{bidirectional pairwise-interaction graph}. To reduce the sensitivity of person-person interaction estimation, we further introduce a unidirectional hyper-interaction graph that models the consensus of interaction between a focal person and any person in a local region around.
			%		\wsss{Finally, we modify the PageRank algorithm to infer the activeness of persons on a Hybrid-Interaction Graph (HIG), the union of the pairwise-interaction and hyper-interaction graphs}, and we call our algorithm the \emph{PersonRank}.
			\whllll{In order to provide publicable datasets for evaluation}, we have contributed a new dataset called Multi-scene Important People Image Dataset and gathered a NCAA Basketball Image Dataset from sports game sequences. We have demonstrated that the proposed PersonRank outperforms related methods clearly and substantially. Our code and datasets are available at \url{https://weihonglee.github.io/Projects/PersonRank.htm}.}

	\end{abstract}
	
	%%%%%%%%% BODY TEXT
	\section{Introduction}
	\vspace{-0.1cm}
	In some \WH{social} \weihong{activities}, not all of the people involved in an event act equally, and always there are important people among them.
	%people focus on important people who play a more important role than others. 
	For example, people pay more attention to the speaker in \whlll{a} \whnew{ceremony} (Figure \ref{fig:ImageDemo}(b)) , the interviewee (\weihong{Figure \ref{fig:ShowOnFirstPage}}) and \WH{the shooter}
%	the player who \whl{shoots} the ball 
	in a basketball game (Figure \ref{fig:ImageDemo}(a)). In these examples, \textbf{the important people} in an image \WH{play} a more important role than the others in \weihong{the} \WH{image} \WH{and are more related to the event presented in the image \cite{Key_ramanathan2015detecting,VIP_solomon2015vip}}.
	Detecting important people can potentially benefit event recognition and event detection \WH{\cite{Key_ramanathan2015detecting}}. 
	In particular, analyzing the action of the \WHnew{important people} \weihong{enables} intelligent system to better understand what has happened. For instance, a \whlll{``shoot''} in a game is closely related to the most important person, who \whlll{shoots} the ball (see Figure \ref{fig:ImageDemo}). 
	
	%As for Human-Computer Interaction, when multi-persons images were captured, how to detect important people who has interactions with the computer is a significant task. 
	%Besides, nowadays, people tend to preserve memories of events such as basketball game, lectures, vacations or speeches by capturing photos and posting them online. 
	%In general, photos captured in different scene such as sport arenas, classroom, market places or other areas typically contain multiple persons interacting with each other. 
	%Most people are doing ``something", but they are not equally important in images.
	%Thus, respect to an image or the main event in an image, people have not equal importance and certain persons are key character and play a more central role.
	%Analysing the importance of persons in images and detecting important people who is more responsible for events is an interesting and significant task in computer vision field and can help computer understand what happern in images.

	%
	%
	%Thus, identifying the importance of each persons and detect important people is an interesting and significant task in computer vision. 
	
	%\newline
	%\textbf{What is key people and importance of people?}
	\begin{figure}[t]
		\begin{center}
			\label{fig:ShowOnFirstPage}
%			\fbox{\rule{0pt}{2in}\rule{0.9\linewidth}{0pt}}
			\includegraphics[width=0.8\linewidth]{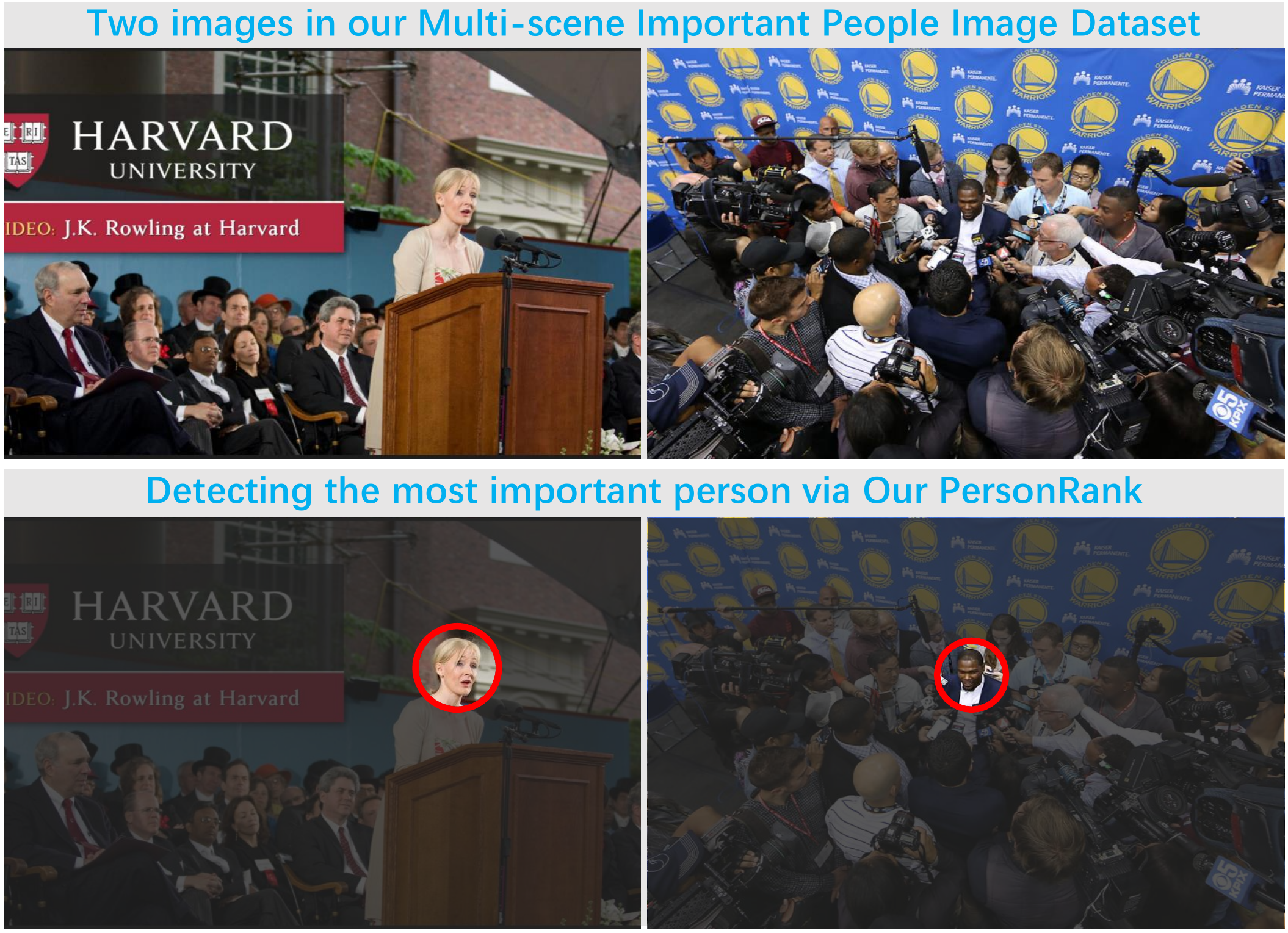}
			\vspace{-0.2cm}
			\centering\small\caption{Important people detection is to detect important people in images who play the central role such as the speaker in a ceremony shown in the figure. Our PersonRank (PR) model is proposed to rank persons in images in terms of importance \WH{scores} and thus find the most important person.}
			\label{fig:ShowOnFirstPage}
		\end{center}
		\vspace{-0.6cm}
	\end{figure}
	
	%\vspace{-0.3cm}
	
	%Nowadays, people tend to preserve events such as basketball game, lectures, vacations or speeches by capturing photos and posting them online. 

	%Besides, in real world, when events occured, not every people involved in events plays an equal role. People pay more attention on important people who plays key role in events.
	
	%Detecting important people is a significant challenge which differentiates Saliency Detection in images.
	
	%Obviously, detecting important person in images can help camera or photographer to automatically detect and track key people and events.
	\WH{While person (face or pedestrian) detection is processed at an ever-faster rate, relatively little work has explored detecting important people against diverse changes of pose, action and appearance of persons and occasions.}
%	Detecting important people in \weihong{images/videos is} a new research topic in computer vision and has been investigated only by a few works recently \whlllll{\cite{VIP_solomon2015vip,Key_ramanathan2015detecting}}. 
%	\whl{Inferring importance of persons and finally detecting the important people \wwhl{from still images} is a challenge.}
\WH{
    Existing work has attempted to use attention model \cite{Key_ramanathan2015detecting} to predict the importance of people directly from their action and appearance or utilize regression model \cite{VIP_solomon2015vip} on spatial and saliency information of face/body to infer relative importance between every two persons. However, these models ignore some semantic information such as interactions between persons, which are essentially useful for inferring the importance of persons.
}
%	Existing work has attempted to use attention model \WH{\cite{Key_ramanathan2015detecting}} or regression model \WH{\cite{VIP_solomon2015vip}} on spatial and appearance information for detecting the \WH{most} important person who \weihong{is} most relevant to an event. 
%	However, these models ignore \WH{some} semantic information such as \weihong{interactions between persons, which are significant for \WH{inferring the importance of persons}.}
	%Moreover, detecting the leader in a violent demonstration who has a great threat for public security is also an important task which will be benefit from important people detection. %It is necessary to detect key people in images.
	
%		\WH{However, it is 
%		identities of persons are unknown and spatial and visual information of them are diverse.
%		We can not simply rank them based on their pose, appearance, action, location and etc. For example, the appearance of the most important people in two images shown in Figure \ref{fig:ShowOnFirstPage} are totally distinct.}
%	\wwhl{
%	Finding out important persons from a large amount of visual cues is not easy for machine, especially when diverse changes of pose, action, appearance of persons are observed at different environments. 
	\WH{Finding important people among individuals in images based on a large amount of visual or spatial cues is challenging, especially when diverse changes of pose, action, appearance of persons are observed at different occasions and identities of these persons are unknown. 
	For example, the appearance of the most important people in two images shown in Figure \ref{fig:ShowOnFirstPage} are totally distinct.}
	
	To \WH{address} this problem\wsss{, we cast the important people detection problem into \WH{ranking nodes} in a graph} \WH{from interactions.}
%	\WHnew{, which models interactions of any pairs of persons and interactions between focal person and any persons in a local region around.}
	Specifically, we propose to consider each person in an image as a node in a graph. Between nodes, we model four types of edge message functions to mimic \WH{interactions} between \wwhl{any pairwise persons}, such as how a person attracts another, how a person \WH{is located} relative to another, how a person's action would affect another and what the appearance difference is between them. This would result in a pairwise-interaction graph which is bidirectional. 
	We also infer the relation between the local consensus of a group of persons and any focal person, which results in a unidirectional hyper-interaction graph. The hyper-interaction graph is unidirectional because the consensus is not physical and thus there is no message \WH{forward from} focal person. The use of hyper-interaction graph can take advantage of local pooling as suggested in other tasks \wwhl{\cite{localpooling_boureau2011ask,localpooling_xiong2017combining,localpooling_yue2015beyond}} so as to \WH{enrich person-person interaction estimation.}
%	alleviate the sensitivity in estimating person-person interactions. 
	Finally, we modify the PageRank algorithm \whll{\cite{PR_leskovec2014mining}} to make it suitable for inferring the activeness of nodes on the union of pairwise-interaction and hyper-interaction graphs, termed the Hybrid-Interaction Graph (HIG), and the most active person is selected as the most important person in an image. We thus call our model the \emph{PersonRank}.

	Since there is \whlll{a} lack of publicable image-based dataset on inferring important people, we also \weihong{contribute} a new dataset, called Multi-scene Important People dataset. This new dataset consists of a large number of images illustrating several events involving multiple persons in different scenes. There are 2310 images mainly from six types of scene. The \weihong{ground-truth} most important person in each image is annotated. We also \weihong{formed} another dataset consisting of images of basketball games \whnew{by} extracting event-starting frame from the NCAA Basketball Dataset \cite{Key_ramanathan2015detecting}. This dataset contains more than 10,000 images and annotations are provided. 
	\whll{Although, there is only one dataset named Image-Level Dataset collected for important people detection before by Solomon et al. \cite{VIP_solomon2015vip},
	unfortunately, this Image-Level Dataset is \weihong{not publicably} available, and it contains only 200 images, much smaller than the ones we collected and \weihong{formed} in this work.}
	%Our dataset is much larger than the Image-Level Dataset in terms of the number of images and annotations for important people detection. 
	
	In summary, \wsss{we make} the following \textbf{contributions}. 
	\wsss{First, we have proposed a PersonRank (PR) that first casts the important people detection problem into ranking nodes in a graph \WH{from interactions}, which is efficient and effective; }
	%And it is also the first time that the idea of PageRank, a powerful method for ranking Internet pages, has been conducted on Computer Vision tasks, to the authors' knowledge.
	Second, we have collected a large images dataset which consists of different events and formed a large basketball \whnew{games} images dataset with \whll{massive}
%	a variety of 
	annotations for important people detection. 
	Experimental results show that our PersonRank (PR) model has obtained the state-of-the-art performance on the two datasets.
	
	\begin{figure*}[t]
		\begin{center}
			\label{fig:FrameWork}
%			\fbox{\rule{0pt}{2in}\rule{0.9\linewidth}{0pt}}
			\includegraphics[width=0.8\linewidth]{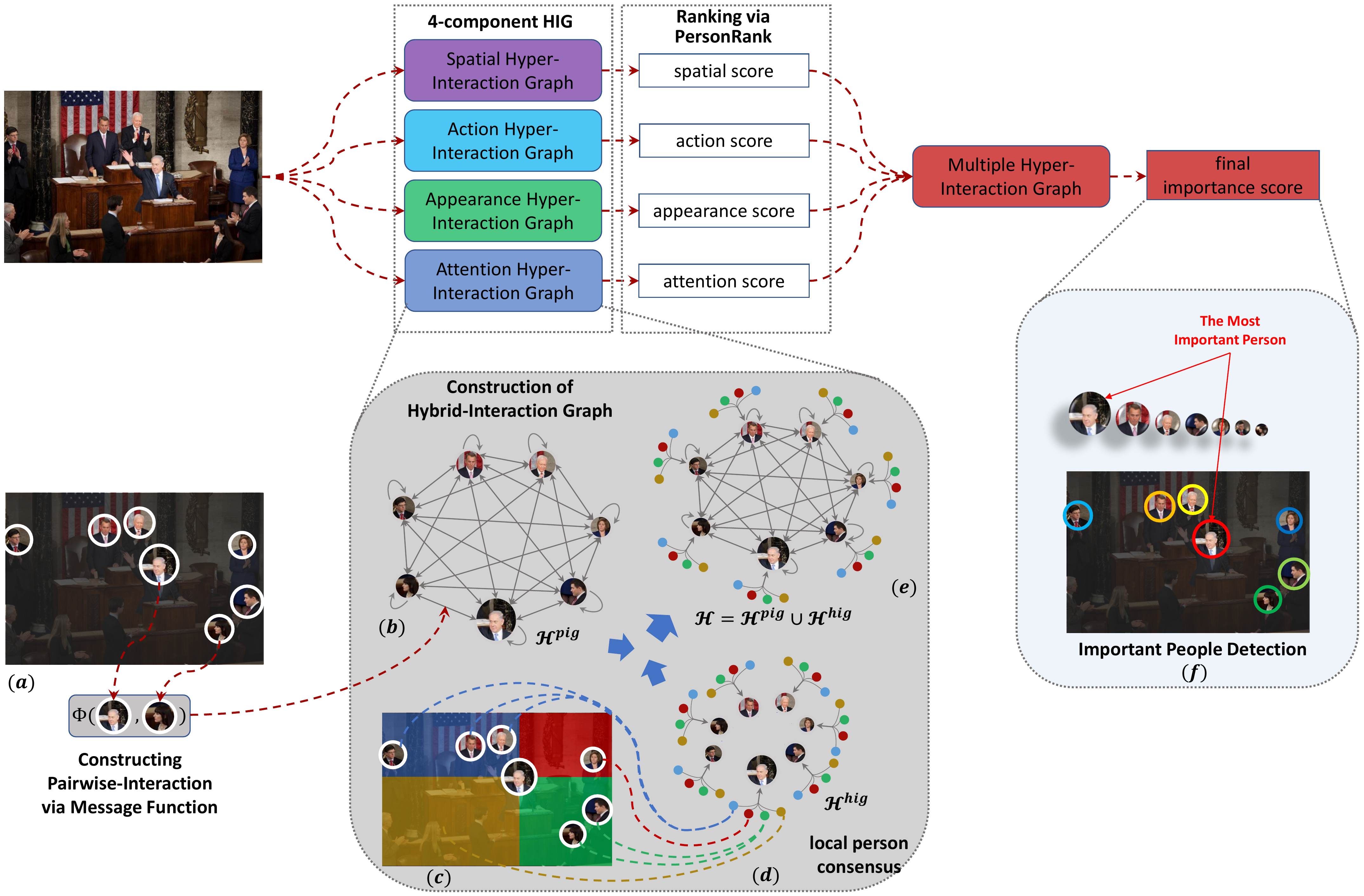}
			\vspace{-0.2cm}
			\centering\small\caption{An illustration of our PersonRank (PR) model. First of all, we detected all persons and initialized them with equal importance score. The \wwhl{Hybrid-Interaction Graph} is constituted for modeling person-person interaction and region-person dependency. Finally, we \weihong{inferred} the importance of persons (Better viewed in color) and the most important person (the person in red box) was selected.}
				%For each person, we extract features and formulated hierarchy interaction graph of these persons by message functions. Finally, we \weihong{inferred} the importance of persons (Better viewed in color) and the most important person (the person in red box) was selected.}
			\label{fig:FrameWork}
		\end{center}
		\vspace{-0.7cm}
	\end{figure*}

	%However, not every people involved in events plays an equal part and people tend to pay more attention on the key people who plays key role in events, in our daily life. 

	%When a Multi-peoson image were presented, there is usually some stories behind the situation that brought them together: a lecture, a speech, a talk or just gathering of a group of friends. 
	%However, not every person in images plays an equal part. 
	%Certain person is the key character and plays more important role (e.g., The speaker of a lecture/speech or a leader of a violent demonstration) called key people in this work. 
	%It is obvious that these key people in images have great impact on Image Understanding (e.g., Event Recognition in Images). 
	%It is proved that detecting the most key people can improve the event recognition performance. 

	\vspace{-0.1cm}
	%-------------------------------------------------------------------------
	\section{Related Work}
	\vspace{-0.1cm}
	\noindent\textbf{Persons and General Object Importance.}
	Recently, several works \cite{IP_berg2012understanding,IP_hwang2012learning,IP_le2007finding,IP_lee2012discovering,IP_spain2011measuring,IP_lee2015predicting,VIP_solomon2015vip} were proposed to study the importance of generic object categories and persons. \weihong{Solomon et al. \cite{VIP_solomon2015vip} studied relative importance between persons and developed regression model for predicting the importance of persons in an image. 
		\whlll{In contrast to using regression model in \cite{VIP_solomon2015vip}, we develop a graph-based model to predict the importance of persons, \WH{where not only  interactions but also importance scores of others are taken into account.}}
%		the interaction between persons as message \whnew{communicated} between nodes in a \whl{graph} and predict the importance of persons from \whl{the graph} by eigen-analysis. 
		In addition, only spatial information was explored in \cite{VIP_solomon2015vip}, and in comparison more significant context \weihong{information} such as attention, action and appearance \weihong{information} are explored }\WH{ and fused effectively in this work}
%		for important people detection in our work.}
	%However, only spatial information is explored in \cite{VIP_solomon2015vip}, while the appearance features, action features, and attention features which are significant cues for important people detection are ignored for context informations were used for modeling interactions between each person. 
	Ramanathan et al. \cite{Key_ramanathan2015detecting} \WH{trained} ``attention" model \WH{with event recognition labels} to detect key actors in basketball games videos. Compared to \cite{Key_ramanathan2015detecting}, our model does not rely on temporal information \WH{and event annotations}, but deeply explores the \wwhl{pairwise-interaction} and \wss{the hyper-interaction between persons} \wsss{directly on images}. 
	%\weihong{Instead, we model the interaction and the message function between persons in order to find the most active one in an image.}
	Some \weihong{researches} also studied detecting \weihong{important people and objects} in egocentric videos and learned a regressor to predict important regions with which the camera wearer has interaction based on cues such as the nearness to hands, gaze, and frequency of occurrence \cite{IP_lee2012discovering,IP_lee2015predicting}. However, these approaches are not suitable for detecting important people \weihong{in} still images.
	%\noindent\textbf{Attention Model.}
	%Recently, ``attention" model \cite{Atten_sankaran2016temporal}, \cite{Atten_wang2016survey}, \cite{Atten_xu2015show} is widely used in computer vision such as language translation, video-caption, image-caption and etc.
	%In general, ``attention" model can be used for concentrating on some certain parts of the information.
	%While for image-caption, Xu et al. \cite{Atten_xu2015show} used ``attention" model to attend to different regions in the image for image caption. 
	%And we solved the learning of models as a two-class classification problems.
	
	\vspace{0.05cm}
	
	\noindent\textbf{Social Interaction.}
	Since we model the interaction between persons in an image, detecting or modeling social interactions \cite{So_fathi2012social,So_amer2015human,So_aghaei2016whom} is related to ours. Fathi et al. \cite{So_fathi2012social} proposed a method for detection and recognition of social interactions in the first-person video of social events. Amer et al. \cite{So_amer2015human} presented an approach to \weihong{model} social interactions on a ``Tower Game" video dataset which is also an \weihong{egocentric videos dataset}. However, these works did not develop for further \weihong{identifying} important people, while we use edge message functions to model the interaction between persons and develop PersonRank model to select the most active person.
	
	\vspace{0.05cm}
	
	\noindent\textbf{Saliency Detection.}
	\weihong{
		Related to important people/objects detection, saliency detection tells where an image draws attention of viewers. It gains huge benefit from the recent large-scale datasets, and a variety of models \cite{Sa_pan2016shallow,Sa1_salMetrics_Bylinskii,Sa1_zhao2015saliency,Sa1_borji2016reconciling} have been developed.
		%Two models using shallow feature or deep feature were proposed for saliency detection in \cite{Sa_pan2016shallow} and can detect saliency area well.
		However, the saliency detection approach may not work well for detecting the most important person if multiple people exist in an image, due to \WH{the} lack of analysis on the relationship between persons. This will be verified by our experiments.
	}
	
	%\noindent\textbf{Person Re-identification.}
	%\whl{
	%Since persons in images are detected first and features of them are extracted. Person Re-Identification, matches images of the same person across camera views, is related to important people detection. However, they are different topics and have different stories. /***weihong: show differences between Person Re-id and Important People Detection***/.
	%	}

	%when a multi-person image which depicts an event was presented, saliency detector cannot analyze the importance of difference persons and detect the important person who plays a key role in the image. Our work focus on using some context informations to analyse interactions of persons in images and detect the important people.

	%Another similar problem is the important person detection in images. In this work, we present a new model and dataset for this specific setting.

	%\textbf{PageRank}:
	
	%The PageRank is a widely used scoring function of networks in general and of the World Wide Web graph in particular.
	
	%\textbf{Multiplex PageRank}\\
	
	\vspace{-0.1cm}
	\section{Approach}
	
	%As far as we know, it is the first time conduct PageRank Model on Computer Vision field. 
	%\whl{
	%Instead of simply inferring important people from person-person interactions, we exploit embedding latent variables which corrrespond to regions to capture the complicated dependencies between regions and persons called region-person interaction.
	%In addition, we design message function for modeling these interactions for important people detection.
	%Instead of simply inferring important people from person-person interactions, we consider mining more global structural interactions between regions and persons. /***weihong:more what information?***/.
	%Therefore, we embed latent nodes and formulating region-person interactions to caputre complicated /***weihong: how to describe it ?***/ for important people detection.
	%modeling regions as latent nodes and formulating region-person interaction to capture complicated 
	%, and we detail the whole idea in the following sections.
	
	%In particular, all detected persons in an images are treated as nodes in a graph, denoted as $\mathcal{V}_{p}$. 
	%Further, as for the $i$ th person $p_i$, the image is split into 4 regions $r_i \in \mathcal{V}_r$ ($r_i = \text{left-top, left-bottom, right-top or right-down}$) are embedded as latent nodes.
	%denoted as $\{\mathcal{V}_{r}\}_{r \in \{\text{left-top, left-bottom, right-top and right-down}\}}$.
	%}
	\vspace{-0.1cm}
	We propose a \wwhl{Hybrid-Interaction Graph (HIG)} based PersonRank (PR) to infer \WH{the} importance of \WH{each person} in this section. In the following, we first present four types of message functions used for modeling the interaction between \wwhl{any pairwise persons}, then detail the construction of the HIG,
%	Hybrid-Interaction Graph, 
	and finally describe the technique to \WH{find important people}. An illustration of our method is shown in Figure \ref{fig:FrameWork}.

	\vspace{-0.15cm}
	\subsection{Modeling Interactions by Message Functions}\label{section:MFunc}
	\vspace{-0.15cm}
	We treat people in an image as nodes in a \whll{graph} and model the interaction between any \wwhl{pairwise} persons as the edge message communicated between them via message functions. We form four types of \wsss{edge} message functions, including spatial and action message functions, appearance message function, and attention message function in order to describe how a person attracts another, how a person \WH{is located} relative to another, how a person's action would affect another and what the appearance difference is between them. 
	\wwhl{Based on each type of message function between nodes (persons), a bidirectional pairwise-interaction graph is generated in the next section.}
%	Based on each type of message function between nodes (persons), a Hyper-Interaction graph is first generated as a bidirectional graph 
	
	%
%	\vspace{0.05cm}
	\noindent\textbf{Spatial and Action Message Functions}.
%	\subsubsection{Spatial and Action Message Functions}
%	\vspace{-0.1cm}
	Given the relative spatial feature $\phi^{s}_{p_i}$ and action feature $\phi^{ac}_{p_i}$ for any person $p_i$, the message function $\varPhi^{s}(\phi^{s}_{p_i},\phi^{s}_{p_j})$ and $\varPhi^{ac}(\phi^{ac}_{p_i},\phi^{ac}_{p_j})$ are used to measure how $p_j$ locates relative to $p_i$ and how the action of $p_j$ would affect $p_i$.
	%the relative importance/***Jason: good word?***/ between $f_i$ and $f_j$.
	%\weihong{to approximate the difference importance of them} ($r_j - r_i$)/***Jason:what?***/. 
	For this purpose, \weihong{the spatial and action message functions} $\varPhi^{s}(\phi^{s}_{p_i},\phi^s_{p_j})$ and $\varPhi^{ac}(\phi^{ac}_{p_i},\phi^{ac}_{p_j})$ are denoted below, respectively:
	\vspace{-0.15cm}
	\begin{equation}
	\small
	\begin{split}
	&\varPhi^{s}(\phi^s_{p_i},\phi^s_{p_j})={\mathbf{w}^{s}}^T exp(\phi^{s}_{p_j}-\phi^{s}_{p_i}),\\
	&\varPhi^{ac}(\phi^{ac}_{p_i},\phi^{ac}_{p_j})={\mathbf{w}^{ac}}^T exp(\phi^{ac}_{p_j}-\phi^{ac}_{p_i}),\\
	\end{split}
	\vspace{-0.15cm}
	\end{equation}
%	where $\phi^{s}_{p_i}$ and $\phi^{ac}_{p_i}$ are the spatial and action feature representation for the $i^{th}$ person, and 
	where $\mathbf{w}^{s}$ and $\mathbf{w}^{ac}$ are learned by a two-class support vector machine in order to identify the difference between an important person and any non-important person from the difference between any two non-important people.
	%weihon¢gnew{so that $\varPhi^{s}(\phi^s_{f_i},\phi^s_{f_*})$ and $\varPhi^{ac}(\phi^{ac}_{f_i},\phi^{ac}_{f_*})$ where $f_*$ is the important people is larger than $\varPhi^{s}(\phi^s_{f_i},\phi^s_{f_k})$ and $\varPhi^{ac}(\phi^{ac}_{f_i},\phi^{ac}_{f_k})$ where $f_k$ is the non-important people}/***Jason: need to explicitly tell what these projections are used***/. 
%	With these two types of message functions, we form a \whll{bidirectional}
%	hierarchy 
%	spatial-interaction based and an action-interaction based graphs.
	
%	\vspace{-0.3cm}
	
%	\subsubsection{Appearance Message Function}
%	\vspace{0.05cm}
	\noindent \textbf{Appearance Message Function}. The appearance message function is to model the difference between the appearance of \wwhl{pairwise} persons, and this is the fact that the appearance of the important people in images could differ from the others (see Figure \ref{fig:ImageDemo}). To this end, given the appearance feature \whlll{$\phi^{ap}_{p_i}$} of the $i^{th}$ person, the message function $\varPhi^{ap}(\phi^{ap}_{p_i},\phi^{ap}_{p_j})$ is designed to measure the appearance difference \whnew{between} \wwhl{pairwise} persons $p_i$ and $p_j$:
	\vspace{-0.15cm}
	\begin{equation}
	\small
	\varPhi^{ap}(\phi^{ap}_{p_i},\phi^{ap}_{p_j}) = {\mathbf{w}^{ap}}^{T}|\phi^{ap}_{p_j}-\phi^{ap}_{p_i}|,
	\vspace{-0.15cm}
	\end{equation}
	where $\mathbf{w}^{ap}$ is also learned by a two-class SVM so that the \whnew{appearance difference} between an important person and any non-important person is larger than the one between any two non-important people.
	
%	\vspace{-0.3cm}
	
%	\subsubsection{Attention Message Function}
%	\vspace{0.05cm}
	\noindent\textbf{Attention Message Function}.
	Since the person who \weihong{attracts} \wsss{others} \whlll{(persons in the image)} mostly
	% in the image 
	is more likely to be the most important person, we form an attention-message function $\varPhi^{at}(\phi^{at}_{p_i}, \phi^{at}_{p_j})(i\neq j)$ where $\phi^{at}_{p_i}$ is the attention feature of the $i^{th}$ person in order to describe the likelihood of the person $p_i$ looking at person $p_j$ as follow \WH{(similar to \cite{So_fathi2012social})}:
	\vspace{-0.15cm}
			\begin{equation}\label{eq:AttenMess}
			\small
			\varPhi^{at}(\phi^{at}_{p_i}, \phi^{at}_{p_j}) = exp\{\mathbf{v}^{T}_{p_i} \cdot g(\mathbf{w}^{at} \odot (\mathbf{f}_{p_j}-\mathbf{f}_{p_i}))-1\},
			\vspace{-0.15cm}
			\end{equation}
	where $\odot$ is the element-wise product, $g(\cdot)$ is used for normalizing the input vector.
	As shown in Figure \ref{fig:3DScene}, in Eq. (\ref{eq:AttenMess}), 
		$\mathbf{v}_{p_i}$ is the direction of the attention of $p_i$, $g({\mathbf{w}^{at}}\odot(\mathbf{f}_{p_j}-\mathbf{f}_{p_i}))$ is computed to approximate the unit orientation from $p_i$ to $p_j$, and $\mathbf{w}^{at} = [1, c]^T$ is tuned on validation data. 
		Hence the inner product between $\mathbf{v}_{p_i}$ and $g({\mathbf{w}^{at}}\odot(\mathbf{f}_{p_j}-\mathbf{f}_{p_i}))$ can describe \WH{the strength of} attention between $p_i$ and $p_j$.

	%In this way, we are able to rank the activeness of persons in the image and finally select the person who acts the most actively in the network as the important person. We call our method the 
	%PersonRank (CPR) method.
	
	\vspace{-0.15cm}
	\subsection{Inferring Importance of Persons from Hybrid-Interaction Graph}\label{section:SecPR}
	\vspace{-0.15cm}
	%For all persons in an image and each kind of feature, we generated a network $\mathbf{G} = (\mathbf{V},\mathbf{E})$ with links $\mathbf{E}_{i,j}=\varPhi(\phi_{f_i},\phi_{f_j})$ which is generated by the message function. Thus, the PersonRank model was conducted on the generated network so that each person can be rank in terms of importance. Finally, the important people in the image can be detected precisely.
	%With the PersonRank being conducted on $\mathbf{G} = (\mathbf{V},\mathbf{E})$ gained before, the importance score $r_{f_i}$ is defined as:
	\whlll{
	Suppose that a set of $N$ persons $\{p_i\}^{N}_{i=1}$
%	$\phi_{p_i}$ 
	are detected in a still image. %,  and called focal persons in this work. %let $\mathcal{V}^p=\{ \phi_{p_1}, \phi_{p_2}, \cdots, \phi_{p_N} \}$.
	In order to find important \wwhl{people}, we form a Hybrid-Interacton Graph (HIG),
	%Hyper-Interaction Graph (HIG), 
	denoted as $\mathcal{H}=(\mathcal{V},\mathcal{E},\mathbf{G})$, where $\mathcal{V} = \mathcal{V}^{p}\cup\mathcal{V}^{r}$ is a set of nodes, $\mathcal{E}= \mathcal{E}^p\cup \mathcal{E}^r$ is a set of edges and $\mathbf{G}$ is \wsss{an} interaction matrix.
%	$\mathcal{V} = \{\mathcal{V}^p,\mathcal{V}^r\}$ is a set of nodes, $\mathcal{E}= \{\mathcal{E}^p,\mathcal{E}^r\}$ is a set of edges and $\mathbf{G}$ is a interaction matrix.
}

%\vspace{0.1cm}

	\noindent \textbf{\wwhl{Constructing Pairwise-Interaction Graph $\mathcal{H}^{pig}=(\mathcal{V}^p, \mathcal{E}^p, \mathbf{G}^p)$}}:
	\whlll{
	First of all, as shown in Figure \WH{\ref{fig:FrameWork}(b)}, we construct a bidirectional graph $\mathcal{H}^{pig}$, where $\mathcal{V}^p=\{\mathbf{V}_{p_i}\}_{i=1}^{N}$ are nodes representing persons $\{{p_i}\}_{i=1}^{N}$ and $\mathcal{E}^p=\{\mathbf{e}^{p}_{1,1},\cdots,\mathbf{e}^{p}_{i,j},\cdots,\mathbf{e}^{p}_{N,N}\}$ are edges.
	\whlllll{Here, each directed edge $\mathbf{e}^{p}_{i,j}=(\mathbf{V}_{p_i},\mathbf{V}_{p_j})$ is used to model the interaction between pairwise persons.
	For each type of feature\footnote{The feature is mentioned in Sec. \ref{section:MFunc} and to be elaborated in Sec. \ref{section:details-impl}} $\phi_{p_i}^z, (z\in\{s,ac,ap,at\})$, 
	we design four types of edge message functions $\varPhi^{z}(\phi^z_{p_i},\phi^z_{p_j}),(z\in\{s,ac,ap,at\})$ formulated in the last section and assign $\mathbf{G}^{p}_{i,j} = \varPhi(\mathbf{e}^{p}_{i,j})=\varPhi^{z}(\phi^z_{p_i},\phi^z_{p_j})$ to each edge in order to capture how a person is interacting with another.}
%	\whllll{For every type of feature\footnote{The feature is mentioned in Sec. \ref{section:MFunc} and to be elaborated in Sec. \ref{section:details-impl})} $\phi_{p_i}^z, z\in\{s,ac,ap,at\}$ , each directed edge $\mathbf{e}^{p}_{i,j}=(\phi^{z}_{p_i},\phi^{z}_{p_j})$ is used to model the interaction between pairwise persons.}
%	\wsss{Each directed edge $\mathbf{e}^{p}_{i,j}=(\phi_{p_i},\phi_{p_j})$ is an ordered pair, and $\phi_{p_i}=[\phi_{p_i}^s; \phi_{p_i}^{ac};\phi_{p_i}^{ap};\phi_{p_i}^{at}]$ denotes the feature 
%		representation of each person $p_i$, where $\phi_{p_i}^z, z=\{s,ac,ap,at\}$ are features used to model the interaction between persons in Sec. \ref{section:MFunc} and will be detailed in Sec. \ref{section:details-impl}.}
	%For capturing person-person interaction among $\phi(p_1), ..., \phi(p_N)$, 
%	For capturing how a person is interacting with another,
%	we design four types of edge message functions $\varPhi^{z}(\phi^z_{p_i},\phi^z_{p_j}),z\in\{s,ac,ap,at\}$
%%	$\varPhi^{s}(\phi^s_{p_i},\phi^s_{p_j}), \varPhi^{ac}(\phi^{ac}_{p_i},\phi^{ac}_{p_j}), \varPhi^{ap}(\phi^{ap}_{p_i},\phi^{ap}_{p_j})$ and $\varPhi^{at}(\phi^{at}_{p_i}, \phi^{at}_{p_j})$ 
%	formulated in the last section and assign $\mathbf{G}^{p}_{i,j} = \varPhi(\mathbf{e}^{p}_{i,j})$ to each edge.
%	${\mathbf{G}^{p}_{i,j}}^{z} = \varPhi^{z}(\mathbf{e}^{p}_{i,j}), z=\{s,ac,ap,at\}$ to each edge.
	Therefore, for each type of edge message function, the person-person interactions among $\mathcal{V}^p$
%	$\mathbf{V}_{p_1},\mathbf{V}_{p_2},\cdots, \mathbf{V}_{p_N}$ 
	can be expressed as a $N \times N$ bidirectional 
%	interaction-message 
	interaction matrix $\mathbf{G}^{p}$. %, of which the nodes correspond to the $N$ persons' variables.
}

%\vspace{0.1cm}

	\noindent \textbf{\whllll{Constructing Hyper-Interaction Graph $\mathcal{H}^{hig}=(\mathcal{V}, \mathcal{E}^r, \mathbf{G}^r)$}}: %$\mathcal{H} = \mathcal{H}^{bi} \cup \mathcal{H}^{uni}$} 
	In order to utilize the local spatial consensus of person \wsss{attractions on a focal person for identifying whether that focal person is important}, we explore a higher level interaction graph for each focal person.
	%by computing the consensus of person-person interaction from persons within a local region. 
	As shown in Figure \WH{\ref{fig:FrameWork}(c)}, for each focal person $p_i$,
	%called focal person in this work, 
	\whllll{we divide the area around him/her into $Q$ block regions and denote each local region as $r_k^i$ corresponding to node $\mathbf{V}_{r_k^i} \in \mathcal{V}^r$ in graph $\mathcal{H}^{hig}$. 
	Then we compute the maximum of the person-person interaction between any person in each block region and focal person. 
	%in that here are 4 further latent variables of each person.
	Specifically, for each type of feature $\phi^z_{p_i}, (z\in\{s,ac,ap,at\})$,
%	focal person $p_i$, 
	we denote the consensus of persons on the focal person $p_i$ in the local region $r_k^i$ as 
	{\small $\gamma^{z}_{r_k^i} = \max(\phi^{z}_{p^{k}_1},\cdots, \phi^{z}_{p_j^{k}})$}, where $p^{k}_j$ is the person in region $r_k^i$ 
	and $\gamma^{z}_{r_k^i}$ is \whlllll{the feature representation} of $\mathbf{V}_{r_k^i}$.}
%	{\small $\gamma^{z}_{r_k^i} = \max(\phi^{z}_{p^{r^i_k}_1},\cdots, \phi^{z}_{p_j^{r^i_k}})$}, where $p^{r^i_k}_j$ is the person in region $r_k^i$ and $\gamma^{z}_{r_k^i}$ is also the variable of $\mathbf{V}_{r_k^i}$.
%	$\gamma^{z}_{r_k^i} = \max(\phi^{z}_{p^{r^i_k}_1},\cdots, \phi^{z}_{p_j^{r^i_k}})$, where $p^{r^i_k}_j$ is the person in region $r_k^i$.
%	we denote the consensus of the $k^{th}$ local region as $\gamma^{z}_{r_k^i} = \max(\phi^{z}_{p^{r^i_k}_1},\cdots, \phi^{z}_{p^{r^i_k}_j})$ 
%	associate with node $\mathbf{V}_{r_k^i} \in \mathcal{V}^r$ in graph $\mathcal{H}^{uni}$, where $\mathcal{V}^r$ is a node set representing regions, is the variable of $\mathbf{V}_{r_k^i}$ and $p^{r^i_k}_j$ is the person in region $r_k^i$.
%	$\gamma^{z}(r_k^i) \in \mathcal{V}_r$, where $\mathcal{V}_r$ is a node set,
	%embeded into $\mathcal{H}^{bi}$, 
%	$\gamma^{z}(r_k^i) = \max(\phi^{z}(p^{r^i_k}_1),\cdots, \phi^{z}(p^{r^i_k}_j))$ and $p^{r^i_k}_j$ is the person in region $r_k^i$. 
	\whlllll{Then, as shown in Figure \WH{\ref{fig:FrameWork}(d)},  the unidirectional hyper-edge $\mathbf{e}^{r}_{i}=(\mathbf{V}_{r_1^i},\cdots,\mathbf{V}_{r_Q^i}, \mathbf{V}_{p_i}) \in \mathcal{E}^r$ %($k=1,\cdots,Q$) 
	between 
%	such a 
	$Q$ consensuses and focal person is formed by message function as similarly described in the last section.
	So that we use a $Q \times N$ unidirectional interaction matrix $\mathbf{G}^r$, with $\mathbf{G}^{r}_{k,i} = \varGamma(\gamma^{z}_{r_k^i}, \phi^{z}_{p_i}) = \mathbf{\delta}^T(\phi^{z}_{p_i} -\gamma^{z}_{r_k^i} )$, $(k=1,\cdots,Q)$, to model such a region-person hyper-interaction and assign $\sum_{k=1}^{Q}\mathbf{G}^{r}_{k,i}$ to $\mathbf{e}^{r}_{i}$.
	Note that we form the unidirectional edge rather than the bidirectional one because each consensus is not physical person and thus there is no backward message sent from the focal person.}
	\whllll{We empirically set\footnote{\whlllll{We find setting $Q$ large would not make benefit clearly, so we set $Q=4$ in this work}} $Q=4$ and the vector $\mathbf{\delta}$ is learned by a two-class SVM to separate the most important and non-important person.}

%\vspace{0.05cm}

%\noindent \whlll{\textbf{Complete Person-Person Interaction Graph $\mathcal{H}$}.
	\noindent \wwhl{\textbf{Constructing Hybrid-Interaction Graph $\mathcal{H}$}:
	With the unidirectional Hyper-interaction graph \whlllll{$\mathcal{H}^{hig}$,}
%	=(\mathcal{V}, \mathcal{E}^{r}, \mathbf{G}^{r})$, 
	the \wwhl{Hybrid-Interaction Graph} (Figure \WH{\ref{fig:FrameWork}(e)})
%	complete person-person interaction graph 
	is finally formed as $\mathcal{H}=\mathcal{H}^{pig} \cup \mathcal{H}^{hig}=(\mathcal{V}, \mathcal{E}, \mathbf{G})$, and the full interaction matrix $\mathbf{G}$ that describes pairwise and hyper person-person interactions is defined as:
	\vspace{-0.2cm}
		\begin{equation}\label{eq:PR}
%		\normalsize
		\small
		\mathbf{G} = 
		\left[
		\begin{matrix}
		\mathbf{G}^{p} & \mathbf{0}\\
		\mathbf{G}^{r} & \mathbf{0}
		\end{matrix}
		\right],
		\vspace{-0.2cm}
		\end{equation}%\label{PR}
	where $\mathbf{0}$ denotes a zero matrix and $\mathbf{G}$ is a $(N + Q) \times (N + Q)$ square matrix.
	}

	\noindent \textbf{Inferring Importance}.
	%With the Hyper-Interaction Graph, 
	%we utilize eigen-analysis to infer activeness of each person.
	%Now, given any type of \whnew{feature} $\phi_{f_i}^z$, $z \in \{s, ac, ap, at\}$, we formally form an interaction network $\mathbf{G} = (\mathbf{V},\mathbf{E}^z)$ (as shown in Figure \ref{fig:FrameWork}) that describes the interaction between persons via message functions, where we define the edges $\mathbf{E}_{i,j}^z=\varPhi(\phi_{f_i}^z,\phi_{f_j}^z)$. 
	%The edges in the network are bidirectional and the message between nodes will pass to both nodes. 
	\wsss{We model the activeness of each person in the graph as the accumulated interactions with the others, because active people in a scene are always likely to interact with other active people more.
	\wwhl{Let the importance score of person $p_i$ be denoted by $\lambda_{p_i}$. Hence, in a graph, we can compute the importance score of a node by the accumulation of message function values from other nodes, i.e.:
		\vspace{-0.25cm}
		\begin{equation}\label{eq:PR}
		\small
		\lambda_{p_i} = \frac{(1-\alpha)}{N} + \alpha \sum_{k=1}^{Q} \mathbf{G}^{r}_{k,i} + \alpha \sum_{j=1}^{N}\mathbf{G}^{p}_{j,i}\frac{\lambda_{p_j}}{C_{p_j}}, (i = 1,2,\cdots,N),
		\vspace{-0.1cm}
		\end{equation}
	where $\frac{(1-\alpha)}{N} + \alpha\sum_{k=1}^{Q} \mathbf{G}^{r}_{k,i}$ is the prior importance score inferred from hyper person-person interaction, $\alpha$ is a damping factor that can be set within the range \whll{$0 < \alpha \leq 1$},}
%	between 0 and 1, 
	\wwhl{and $C_{p_j}=\sum_{n=1}^{N}\mathbf{G}^{p}_{j,n}$} is the total number of outbound links from node \whll{$\mathbf{V}_{p_j}$.}
	\wsss{The interaction $\mathbf{G}^{p}_{j,i}$ is weighted by the important score of another person $p_j$, because the more interaction between two persons the more they would communicate and thus the message input from more active people would be more important to estimate the importance (activeness) of a person.}}
		
	We actually modify the PageRank algorithm by adding $\sum_{k=1}^{Q} \mathbf{G}^{r}_{k,i}$ in Eq. (\ref{eq:PR}) to rank the nodes in a graph. The term $\sum_{k=1}^{Q} \mathbf{G}^{r}_{k,i}$ is a specific bias to person $p_i$. In conventional PageRank, the prior score is fixed as $\frac{(1-\alpha)}{N}$ for each person, while in our modification, the prior score is $\frac{(1-\alpha)}{N}+\alpha \sum_{k=1}^{Q} \mathbf{G}^{r}_{k,i}$ and this score differs for different persons $p_i$. Our experiments have shown that such an asymmetric formulation would gain benefit.
%	to person $p_i$, 
%	and $N$ is the total number of persons in an image.

    \wwhl{For computation, we infer the importance of each person in an image by applying eigen-analysis \cite{PR_leskovec2014mining} on the Hybrid-Interaction Graph,
%    the union of bidirectional pairwise person-person interaction graph and the unibirectional hyper one, 
    since computing the important score vector $\mathbf{\Lambda} = [\lambda_{p_1}, \lambda_{p_2}, \cdots \lambda_{p_N}]^{T}$ can be viewed as an eigenvector problem, 
	$\mathbf{\Lambda} = \overline{\mathbf{G}}\mathbf{\Lambda}$, where $\overline{\mathbf{G}} = {\alpha\hat{\mathbf{G}}^p}^T + {\alpha\mathbf{G}^{r}}^T\mathbf{1}\mathbf{1}^T+ \frac{(1-\alpha)\mathbf{1}\mathbf{1}^{T}}{N}$. Here, $\hat{\mathbf{G}}^p_{i,j}=\mathbf{G}^p_{i,j}/\sum_{n=1}^{N}\mathbf{G}^p_{i,n}$ is the element of $\hat{\mathbf{G}}^p$ and $\mathbf{1}$ is an all-ones vector.}

	%Although the formula Eq. (\ref{eq:PR}) has a similar form as PageRank \cite{PR_leskovec2014mining}, the problem and motivation are different.
	
	%The $k$ th region of person $p_i$ is denoted as $R_i$ and the region-person interactions are also formulated by message function .
	
%	\vspace{0.05cm}
	
	\noindent \textbf{Inferring Importance on Multiple Interaction Graphs}.
	Since we have designed four types of edge message functions, four \wwhl{Hybrid-Interaction Graphs} \whlll{$\mathcal{H}^{z},(z \in \{s, ac, ap, at\})$} are formed, and we resort to fuse the importance scores \weihongnew{ $\phi^{\lambda}_{p_i} =[\lambda^{s}_{p_i},\lambda^{ac}_{p_i},\lambda^{ap}_{p_i},\lambda^{at}_{p_i}]^T$ generated by the four graphs for each person}.  For this purpose, 
	we compute \WH{the} fused importance score $R_{p_i}$ \WH{for each person} $p_i$ by 
	\vspace{-0.15cm}
		\begin{equation}
		\small
		R_{p_i} = \prod_{z \in \{s, ac, ap, at\}}{(\lambda^{z}_{p_i})}^{q_z},
		\vspace{-0.2cm}
		\end{equation}
	where $\mathbf{q} = [q_{s}, q_{ac}, q_{ap}, q_{at}]^T$ is a weight vector consists of weight for each graph. Therefore, the fused message function $\varPhi^{f}(R_{p_i},R_{p_j})$ is denoted as:
		\vspace{-0.2cm}
		\begin{equation}
		\small
		\begin{split}
		\varPhi^{f}(R_{p_i},R_{p_j})&= log(\frac{R_{p_j}}{R_{p_i}})= {\mathbf{q}}^T[log(\phi^{\lambda}_{p_j})-log(\phi^{\lambda}_{p_i})].
		\end{split}
		\vspace{-0.2cm}
		\end{equation}
	%\whlll{Therefore, instead of using Multiplex PageRank for inferring importance from multiple graphs, we form a fused Hyper-Interaction Graph $\mathcal{H}^{f} = (\mathcal{V},\mathcal{E},\mathbf{G})$.
%	\whlll{And then we form a fused Hyper-Interaction Graph $\mathcal{H}^{f} = (\mathcal{V},\mathcal{E},\mathbf{G})$
	%, where $\mathbf{E}_{i,j}=\varPhi^{f}(R_{p_i},R_{p_j})$. 
	So eigen-analysis is used again to infer the importance score of each node, and finally the node corresponding to the largest score is selected as the most important person.
	
	%We finally infer the fusion importance score of persons in images.
	Finally, we discuss how to learn $\mathbf{q}$ in the above model.
	Suppose \WH{we are} given a set of persons $\mathbf{p}=\{p_i\}_{i=1}^{N}$
%	$\mathbf{p} = \{p_i|p_i \text{ denotes a person in an image}, i = 1, 2, \cdots N\}$ 
	detected in an image, and \whlllll{we denote $p^{*} \in \mathbf{p}$ as the most important person in the image}. 
	For convenience, we rewrite the fusion message function $\varPhi^{f}(\phi^{\lambda}_{p_i},\phi^{\lambda}_{p_j}) = \WHnew{{\mathbf{q}}^T\mathbf{x}_{i,j}}$, where $\mathbf{x}_{i,j}=log(\phi^{\lambda}_{p_j})-log(\phi^{\lambda}_{p_i})$ is a logarithm difference between $p_i$ and $p_j$. 
%	importance
	The weight vector $\mathbf{q}$ is learned such that for any non-important person $p_i \in \mathbf{p}$, $p_i \neq p^*$, the \WH{distance} between an important person and any non-important person is larger than the one between any two non-important people, that is to learn the projection $\mathbf{q}$ such that
	\vspace{-0.15cm}
		\begin{equation}\label{svmlearnX}
		\small
		\mathbf{q}^T\mathbf{x}_{i,*} >{\mathbf{q}}^T \mathbf{x}_{i,j},
		\vspace{-0.15cm}
		\end{equation}%\label{svmlearnX}
	where $\mathbf{x}_{i,*} = log(\phi^{\lambda}_{p_*})-log(\phi^{\lambda}_{p_i})$.
	
	We optimize the above constraint in a two-class SVM framework implemented by the toolbox in \cite{SVM_tool}. The parameter $\alpha$ in Eq. (\ref{eq:PR}) is set to be 0.85 as a default value.
	All of hyper-parameters of SVM were chosen by cross-validation on the validation set of datasets.

		\begin{figure*}[t]
			\begin{center}
				\label{fig:ImageDemo}
%				\fbox{\rule{0pt}{2in}\rule{0.9\linewidth}{0pt}}
				\includegraphics[width=0.8\linewidth]{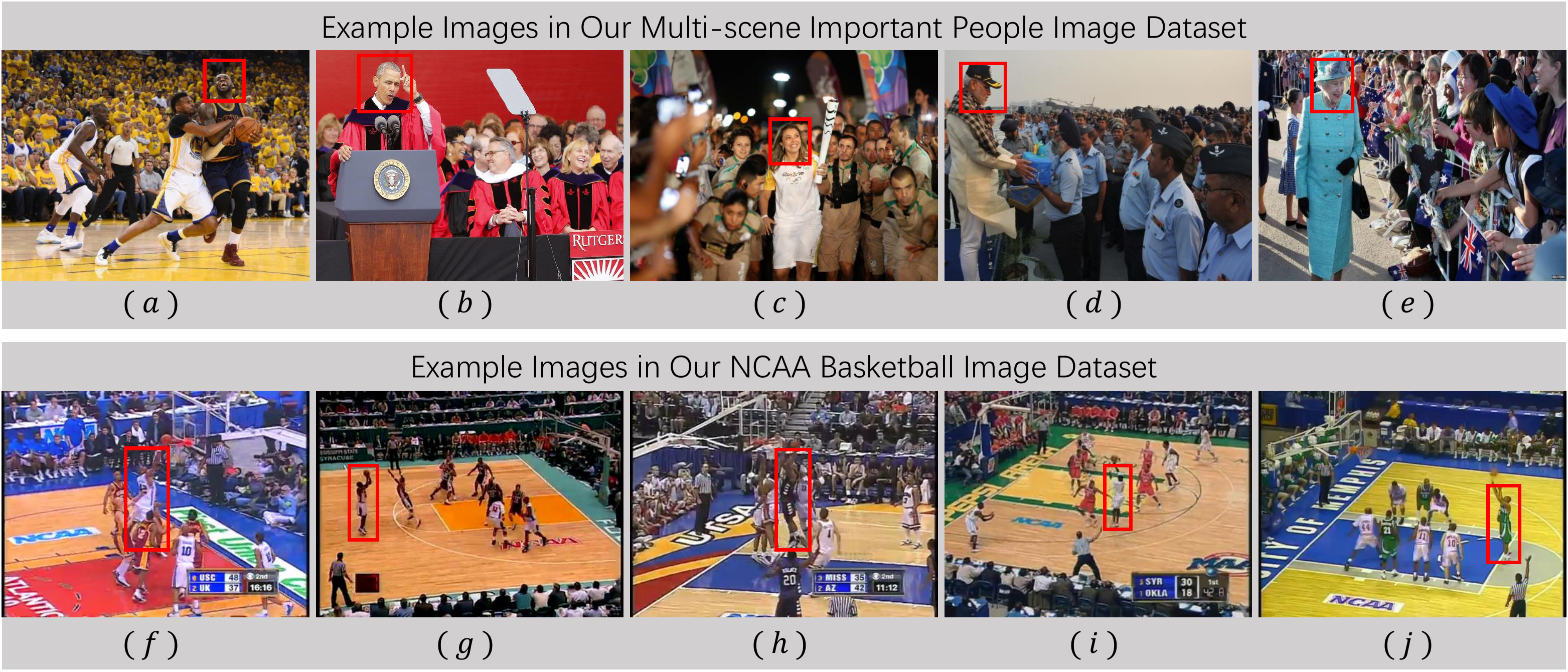}
				\vspace{-0.2cm}
				\centering\small\caption{Some examples in our two new image datasets were shown above. Images shown in first row \weihong{are} examples in Multi-scene Important People Image Dataset while the second row of images are examples in NCAA Basketball Image Dataset. The most important person in each image was annotated in terms of \weihong{face or} body bounding box in Multi-scene Important Persons Image Dataset and NCAA Basketball Image Dataset, respectively. For instance, the person in red bounding box was annotated as the most important people here.}
				\label{fig:ImageDemo}
			\end{center}
			\vspace{-0.75cm}
		\end{figure*}
	\vspace{-0.15cm}
	\subsection{Implementation Details: Importance Feature}\label{section:details-impl}
	\vspace{-0.15cm}
%	\whl{
		%The persons $\{p_i\}_{i=1}^N$ in a given image are first detected whos bounding box is provided and denoted as as $[x_{p_i}, y_{p_i},w_{p_i}, h_{p_i}]$, where $[x_{p_i}, y_{p_i}]$ is the location of the bounding box in the image and $[w_{p_i}, h_{p_i}]$ is the scale of person $p_i$. %Besides, the face orientation $\theta_{p_i} = yaw$
		For convenience of description, we denote the bounding box of a person as $[x_{p_i}, y_{p_i},w_{p_i}, h_{p_i}]$, where $[x_{p_i}, y_{p_i}]$ is the location of $p_i$ in the image and $[w_{p_i}, h_{p_i}]$ is the scale of $p_i$.
		Afterwards, for describing different complicated interactions \wsss{among persons$\{p_i\}^{N}_{i=1}$}, four types of importance features are extracted \wsss{for each person $p_i$: 1) spatial feature ($\phi^{s}_{p_i}$), 2) attention feature ($\phi^{at}_{p_i}$), 3) action feature ($\phi^{ac}_{p_i}$), and 4) appearance feature ($\phi^{ap}_{p_i}$ ).}		%For describing the interaction between persons, four types of person-person features are extracted: 1) spatial feature; 2) action feature; 3) appearance feature; 4) attention feature.

			\begin{figure}[t]
				\begin{center}
					\label{fig:3DScene}
%					\fbox{\rule{0pt}{2in}\rule{0.9\linewidth}{0pt}}
					\includegraphics[width=0.9\linewidth]{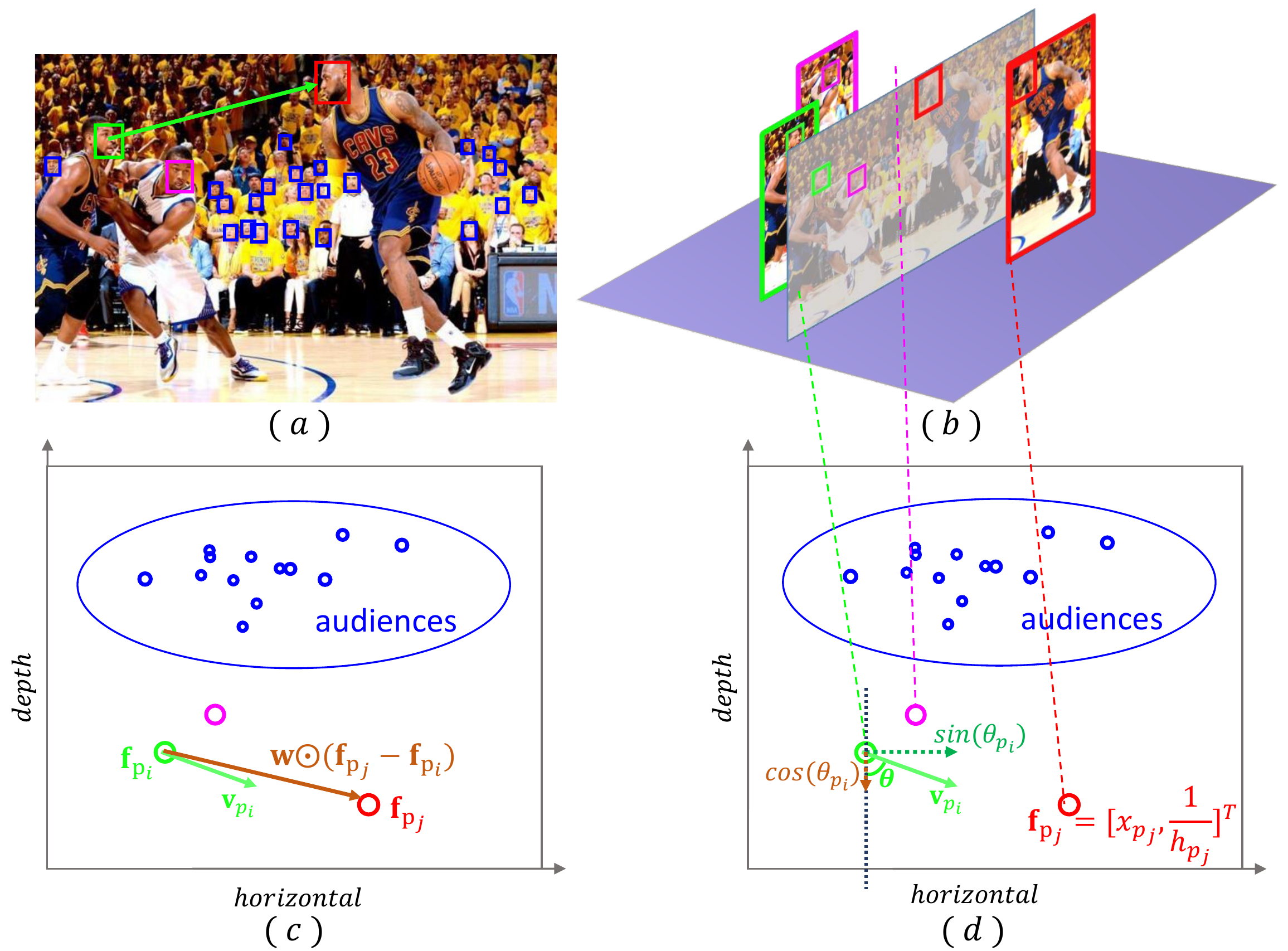}
					\vspace{-0.2cm}
					\centering\small\caption{\whnew{For each input image (a), we use the extracted attention feature to estimate the likelihood of a person looking at another. For convenience of description, we show the bird's eye view of the location and orientation of persons in 3D in (c) and (d). In (c) and (d), $\mathbf{v}_{p_i} = [sin(\theta_{p_i}), cos(\theta_{p_i})]^{T}$ is the direction of the attention of person $p_i$, and $\mathbf{w}^{at} \odot (\mathbf{f}_{p_j} - \mathbf{f}_{p_i})$ (see definition in Eq. (\ref{eq:AttenMess})) is used to approximate the orientation from $p_i$ to $p_j$. The inner product between $\mathbf{v}_{p_i}$ and the normalized $\mathbf{w}^{at} \odot (\mathbf{f}_{p_j} - \mathbf{f}_{p_i})$ can describe how well the attention of $p_i$ is on $p_j$. (Better viewed in color.)}}
					\vspace{-0.8cm}
					\label{fig:3DScene}
				\end{center}
			\end{figure}
	
%	\vspace{0.05cm}
	\noindent \textbf{Spatial Feature}.
	%The spatial feature is to represent where each person locates and how it locates relative to the others. 
	From the perspective of the \wsss{photographers}, they often use a narrow depth-of-field to keep the important people in focus (see Figure \ref{fig:ImageDemo}(c)) so that the important people in photos \wsss{are} likely to be larger, clearer, \weihong{or} closer to image center. Also, the density of persons is also useful; for instance, in the basketball games shown in Figure \ref{fig:ImageDemo}(a), the players (especially the one holding the basketball) are more important than audiences, and thus the density \wsss{helps} distinguish the players from audiences.
	These spatial information is then represented by a 7-dimensional spatial feature vector $\phi^{s}_{p_i} = [\mathbf{s}^{T}_{p_i}, \mathbf{\ell}^{T}_{p_i}, d_{p_i}]^{T}$ for each person $p_i$, including three parts: saliency features ($\mathbf{s}_{p_i}$), location ($\mathbf{\ell}_{p_i}$), and density ($d_{p_i}$).
	
	The saliency feature $\mathbf{s}^{T}_{p_i}$ {is a four dimensional vector} consists of the area of \whll{face/body} bounding box, face sharpness by applying a Sobel filter \cite{Sharpness_sobel2014history} and computing the sum of the gradient energy in a face bounding box, face aspect ratio \cite{VIP_solomon2015vip} and detection confidence computed by \cite{FaceDetector}.
%	 \weihong{are} used as features denoted as a four dimensional vector $\mathbf{s}_{p_i}$. 
	 On the location feature $\mathbf{\ell}^{T}_{p_i}$, we compute the distance between the bounding box center and the image center and compute the distance between each person and the group centroid, the averaged centers of bounding boxes. These distances are normalized by the scale of image. The density of persons $d_{p_i}$ is computed
	within a $m \times m$ support area centered around person $p_i$, where $m$ is one tenth of the width of the image.

	\noindent \textbf{Attention Feature}.
	Based on the location, scale and orientation $\theta_{p_i}$, the attention feature of person $p_i$ is a four dimensional feature denoted as $\phi^{at}_{p_i}=[\mathbf{f}_{p_i}^{T}, \mathbf{v}_{p_i}^{T}]^T$, where $\mathbf{f}_{p_i} = [x_{p_i},\frac{1}{h_{p_i}}]^{T}$ and
	$\mathbf{v}_{p_i} = [sin(\theta_{p_i}),cos(\theta_{p_i})]^{T}$.
	In $\mathbf{f}_{p_i}$, $x_{p_i}$ estimates the horizontal coordinate of $p_i$ in 3D, and $\frac{1}{h_{p_i}}$, the inverse of the height of face bounding box can reflect the depth of $p_i$ from the camera; in $\mathbf{v}_{p_i}$, $\theta_{p_i}$ is the $yaw$ angle of person $p_i$ extracted by the face detector \cite{FaceDetector}, so $sin(\theta_{p_i})$ is the horizontal component and $cos(\theta_{f_i})$ is the depth component of $yaw$ (Figure \ref{fig:3DScene}).
	
%	\vspace{0.1cm}
	\vspace{0.05cm}
	\noindent \textbf{Action Feature}. 
	As shown in Figure \ref{fig:ImageDemo}, identifying action information of each person \weihong{helps detect important people in images}.
	\whnew{For instance, the most important person of ceremony is the speaker who raised his left hand (see Figure \ref{fig:ImageDemo}(b)).}
	%For instance, the most important person in a basketball game is the ``shooter" who shots the ball (see Figure \ref{fig:ImageDemo}(a).
	Thus, for each person $p_i$ in an image, a 2048 dimensional vector $\phi^{ac}_{p_i}$ extracted by a ResNet \weihong{\cite{UCF_feichtenhofer2016convolutional}} on the support region centered at the face of $p_i$ which is 6 times larger than the scale of the face bounding box \whll{or 3 times larger than the body bounding box}, where the network was trained on UCF-101 dataset \cite{UCF101_soomro2012ucf101}.
	
%	\vspace{0.05cm}
	
	\noindent \textbf{Appearance Feature}. 
	Sometimes, the appearance of important people in images, such as the color of clothes, the object \weihong{he/she} operates (e.g. a ball, a microphone), is likely to differ from the others. As shown in Figure \ref{fig:ImageDemo}, the most important person in Figure \ref{fig:ImageDemo} (c) is the torchbearer who holds a torch and wears clothes different from the others.
	We use a ResNet to extract a 2048 dimensional vector around the \whnew{location} of person $p_i$, denoted as $\phi^{ap}_{p_i}$, where the network is pre-trained on ImageNet \cite{Img_he2015deep}.

	\vspace{-0.1cm}
	
	\section{New Important People Detection Image Datasets}
	
	\vspace{-0.15cm}

	%Recently, for important people detection, there is only one small dataset (Image-Level dataset) containing 200 images which was collected by Solomon et al. \cite{VIP_solomon2015vip}. However, the dataset is too small and still not available. Thus, in order to solve this significant detection task, 
	
	We have collected/generated two large datasets: 1) Multi-scene Important People Image Dataset 2) NCAA Basketball Image Dataset\footnote{These two datasets will be \weihong{publicably} available.}. %Most images in the first dataset are close shots while the ones in the second dataset are long shots /***Jason: what?***/.
	
	\vspace{0.05cm}
	\noindent \textbf{1) Multi-scene Important People Image Dataset}.
	The multi-scene image dataset contains 2310 images from more than six types of scene. These images were mined \whll{from Internet}
%	from Filckr 
	using search queries such as ``graduation ceremony", ``people+events", and etc. We manually identified \WHnew{mainly} 6 key scene types listed in Table \ref{tab:Multi-scene}. This dataset includes three subsets: a training set consists of 924 \whnew{images}, a validation set consists of 232 \whnew{images}, and  a testing set consists of 1154 images. 
	\whll{Since faces of most persons are detectable, a robust and stable face detector \cite{FaceDetector} was used to detect persons in all images in our dataset, and the detector has a fairly low false positive rate.}
	%For annotation, a robust and stable face detector \cite{FaceDetector} was used to run though all images in our dataset, and the detector has a fairly low false positive rate. For the training images, some missing faces were then annotated manually. 
	In order to label the most important person in an image, 
	\weihong{ten annotators were asked to vote and the person who is with \whnew{the largest number of votes} is then annotated as the ground-truth most important person in an image.}
	%For instances, the annotators were ask to annotate the speaker in an image for a speech.
	%And the annotators were asked to annotate the important person in images. For instances, the annotators were ask to annotate the speaker in an image for a speech.
	
	\begin{table}[t]
		\footnotesize
		\centering
		\caption{Multi-scene Important People Image Dataset}
		\vspace{-0.2cm}
		\footnotesize
		\resizebox{!}{1.5cm}
		{
		\begin{tabular}{c|c|c}
			%\hline
			\hline
			Scene                          & \# Training (Testing) Images  & Avg. \# persons\\
			\hline
			\hline
			Lecture / Speech         &     360(360)               & 9.45\\
			Demonstration               &     200(200)               & 10.57\\
			Interview                      &     208(207)               & 6.40\\
			Sports                         &     121(122)                 & 8.35\\
			Military                        &     153(152)                 & 8.45\\
			Meeting                      &     38(45)                  & 5.48\\      
			Others                        &     78(68)                  & 7.57\\   
			\hline                 
			\hline
			Total                            &     1156(1154)              &8.59\\
			\hline
			%\hline
		\end{tabular}\label{tab:Multi-scene}
	}
		\vspace{-0.6cm}
	\end{table}
	
%	\vspace{0.05cm}
	
	\noindent \textbf{2) NCAA Basketball Image Dataset}.
	Another natural choice for collecting important people detection images dataset \whnew{is} team sports. In this work, we gathered an image dataset for important people detection by extracting \weihong{frames} which contain ball location annotations from a multi-person event detection video dataset  \cite{Key_ramanathan2015detecting}, covering 10 different types of events. 
	\whll{The bounding box annotations for NCAA basketball dataset are the same as the ones in \cite{Key_ramanathan2015detecting}, where both annotated the shooter as the most important person in a basketball game image. Since faces of the sportsmen are always very small, the whole person body was detected \whlll{by a Multibox detector \cite{Mulbox_szegedy2014scalable,Key_ramanathan2015detecting}}.% /***Jason: detection ?***/
		}
%	The bounding box annotations for NCAA basketball dataset are the same as the ones in \cite{Key_ramanathan2015detecting}, where both annotated the player who was closest (in image space) to the ball as the most important person in a basketball game image.

	\vspace{-0.1cm}
	
	\section{Experiment}
	\vspace{-0.1cm}
	In this section, we conducted evaluations of important people detection on the two new datasets.
%	 \weihong{A demo on two clips of television shows named ``Friends" and a volleyball teaching video is also submitted as the supplementary.}
	\vspace{-0.15cm}
	\subsection{Compared Methods}
	\vspace{-0.2cm}
	\noindent \textbf{Baseline}. We compared our model with several baselines: ``Most-Center'', ``Max-Scale'', ``Max-Face'', ``Max-Pedestrian'' \cite{PD_nam2014local} and ``SVR-Person", where ``Most-Center'' means selecting the person who is closest to the center of an image, ``Max-Scale'' means selecting the person of \whnew{whom the} face/body size is the largest \whnew{one} in an image, ``Max-Face'' means selecting the person of whom the detected face is most confident in an image by using \cite{FaceDetector}, ``Max-Pedestrian'' means selecting the person of whom the pedestrian detection score is the most confident using \weihongnew{\cite{PD_nam2014local}}, \weihongnew{and ``SVR-Person" means using $\nu\text{-SVR}$ \cite{SVM_tool,VIP_solomon2015vip} to predict relative importance between persons and select the most important person based on the concatenatation of the four types of features we used}.
	
%	\vspace{0.1cm}

	\noindent \textbf{``Max-Saliency''}. In order to measure how well a saliency detector performs on important people detection, a recently proposed saliency detector \cite{Sa_pan2016shallow} was also compared. In particular, we implemented the saliency detector to produce saliency map and computed the fraction of saliency intensities inside each face bounding box as a measure of its importance. We denote this compared model \whnew{as} ``Max-Saliency''.
	
%	\vspace{0.1cm}
	
	\noindent \textbf{VIP}\footnote{\WH{We re-implemented VIP as precisely as possible, because the authors' code (including the datasets) in \cite{VIP_solomon2015vip} are not available at our submission time. VIP was trained using the same data set used by other compared methods.}}. We compared the VIP model a recent important people detection model proposed by Solomon et al. \cite{VIP_solomon2015vip}.

	\vspace{-0.15cm}
	\subsection{Evaluation Criterion}
	\vspace{-0.2cm}
	\whll{
		For quantifying the performance of different methods on important people detection, the mean Average Precision (mAP) \wsss{\cite{ObjectDetect_liu2016ssd,ObjectDetect_redmon2016you}}, a criterion which is widely used in object detection, is utilized for judging the correctness of detected important people and reported in the table. 
		%Averaged precision (AP) is calculated for each test image and then we average the APs across all the test images to get the final result.
		In addition, the \whllll{cumulative} matching characteristics (CMC) curve is plotted to show the results of top $k$-rank important people.
		}

%	For quantifying the performance of different methods on important people detection, the cumulated matching characteristics (CMC) curve is plotted to show the results of top $k$-rank important people.
%	\WHnew{The rank-1 matching rate which tells how well the estimated the most important person match the ground-truth one in an image is also reported in the table.}

	%In addition, 
	%we report the precision results which tell how well the estimated the most important person match the ground-truth one in an image. 
	%In addition, the cumulated matching characteristics (CMC) curve is also plotted to show the results of top $k$-rank important people.
	\vspace{-0.15cm}
	\subsection{Experiments on Multi-scene Important People Image Dataset}
%	\vspace{-0.5cm}
\vspace{-0.2cm}
	\whll{All methods were conducted on Multi-scene Important People Dataset and the mean Average Precision for important people detection were shown in Table \ref{tab:comBaseMulti}.
		%Table \ref{tab:comBaseMulti} shows the results for different methods on Multi-scene Important People Detection Dataset. 
		The best baseline achieved 75.9\% mAP and VIP obtained 76.1\%, and our approach achieved 88.6\%.
		Overall, we achieved improvement 14.2\% and 14.0\% over the ``SVR-Person'' baseline and VIP, respectively.
		%Compared to these two regression models for important people detection, 
		The result shows the advantages and effectiveness of our approach.}

	\begin{table}[tp]
		\vspace{-0.2cm}
		\centering
		\footnotesize
		\caption{Mean Average Precison (\%) for Evaluation of Different Methods on Multi-scene Important People Image Dataset}
		\vspace{-0.2cm}
		\resizebox{!}{1.55cm}
		{
		\scriptsize
		\begin{tabular}{c|c|c|c|c|c|c|c|c|c}
			\hline
			\multirow{2}[2]{*}{Method} & Max-  & Max-  & Max-  & Most- & Max-  & SVR-  & \multirow{1}[2]{*}{VIP} & \multirow{1}[2]{*}{\textbf{Ours}}&\multirow{1}[2]{*}{\textbf{Ours}} \\
			& Face  & Pedestrian & Saliency & Center & Scale & Person &   \multirow{1}[2]{*}{\cite{VIP_solomon2015vip}}    &\multirow{1}[2]{*}{($\textbf{PR}^{\text{pig}}$)} &\multirow{1}[2]{*}{($\textbf{PR}$)}  \\
			\hline
			\hline
			Lecture/Speech & 36.4 & 28.2 & 38.3 & 39.4 & 77.8 & \color{blue}{79.9} & 69.6 & {\seccolor{89.5}} & {\bf \color{red}{90.2}} \\
			Demonstration & 29.9 & 27.2 & 45.4 & 59.0 & 75.3 & 77.5 & \color{blue}{84.3} & {\seccolor{90.8}}& {\bf \color{red}{92.0}} \\
			Interview & 36.9 & 36.6 & 36.8 & 59.6 & 78.5 & 77.7 & \color{blue}{85.0} & {\seccolor{89.5}} & {\bf \color{red}{90.2}} \\
			Sports & 35.8 & 33.8 & 40.1 & 60.9 & 67.4 & 69.0 & \color{blue}{79.5} & {\seccolor{80.5}} & {\bf \color{red}{83.6}}\\
			military & 37.9 & 28.5 & 43.3 & 42.3 & 62.6 & \color{blue}{75.4} & 67.7& {\seccolor{85.1}} & {\bf \color{red}{86.5}}\\
			Meeting & 43.2 & 36.2 & 45.1 & 58.6 & \color{blue}{69.0} & 57.5 & 67.9 & {\seccolor{75.2}} & {\bf \color{red}{76.5}}\\
			Other & 35.3 & 31.4 & 37.5 & 57.9 & 74.5 & 69.6 & \color{blue}{76.8} & {\seccolor{86.7}} & {\bf \color{red}{86.7}}\\
			\hline
			\hline
			Total & 35.7 & 30.7 & 40.3 & 50.9 & 73.9 & 75.9 & \color{blue}{76.1} & {\seccolor{87.5}} & {\bf \color{red}{88.6}}\\
			\hline
		\end{tabular}%
	}
		\label{tab:comBaseMulti}%
		\vspace{-0.45cm}
	\end{table}%

\begin{figure}[tp]
		%\label{fig:time_methods}
		\begin{center}
			{\scriptsize
				\subfigure[{\scriptsize Multi-scene Important People Image Dataset}] % caption for subfigure
				{
					\label{fig:CMCcurve}
%					\fbox{\rule{0pt}{2in}\rule{0.9\linewidth}{0pt}}
					\includegraphics[width=0.48\linewidth]{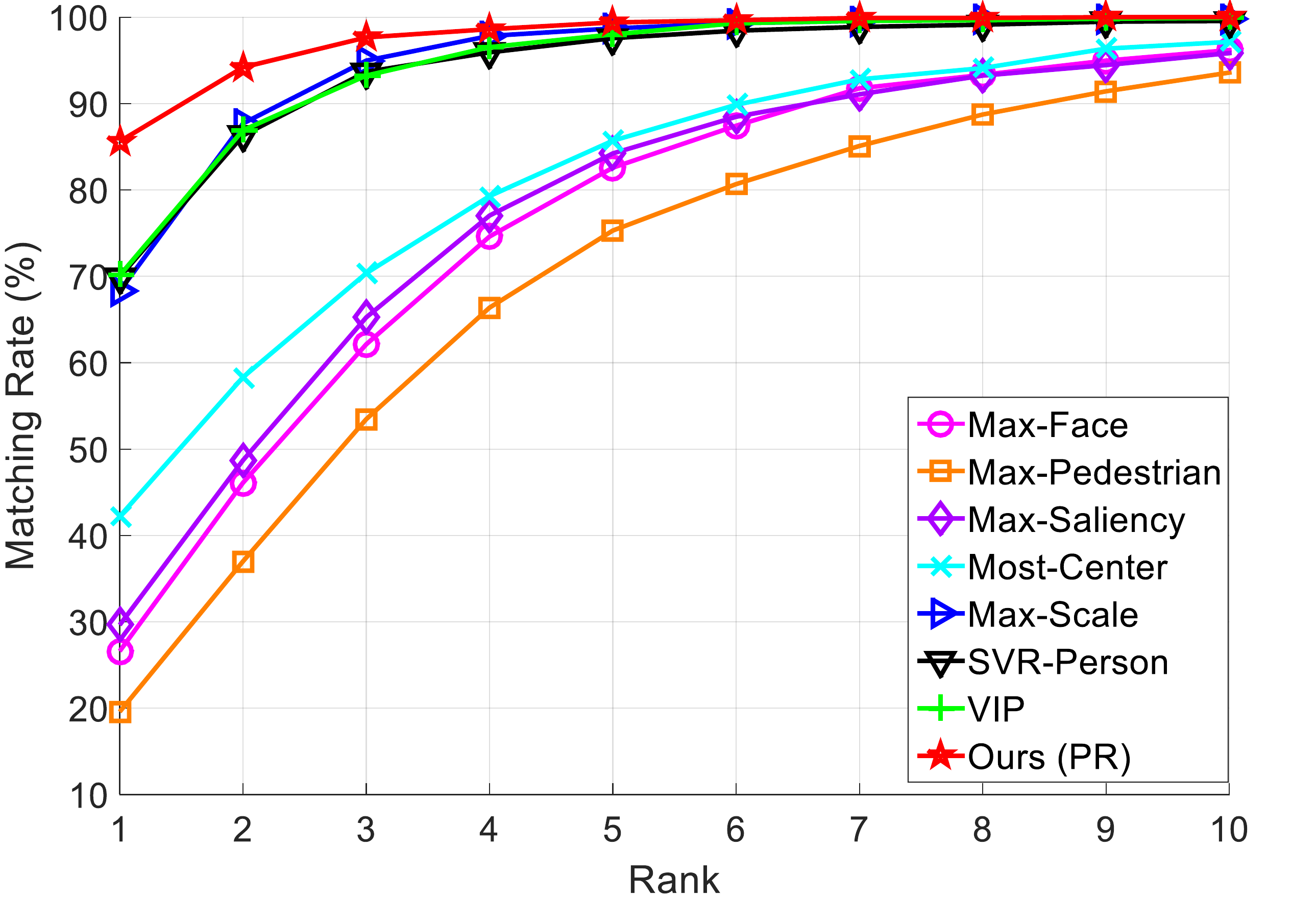}
				}
				\hskip -0.3cm
				\subfigure[{\scriptsize NCAA Basketball Image Dataset}] % caption for subfigure
				{
					\label{fig:NCAACMC}
%					\fbox{\rule{0pt}{2in}\rule{0.9\linewidth}{0pt}}
					\includegraphics[width=0.48\linewidth]{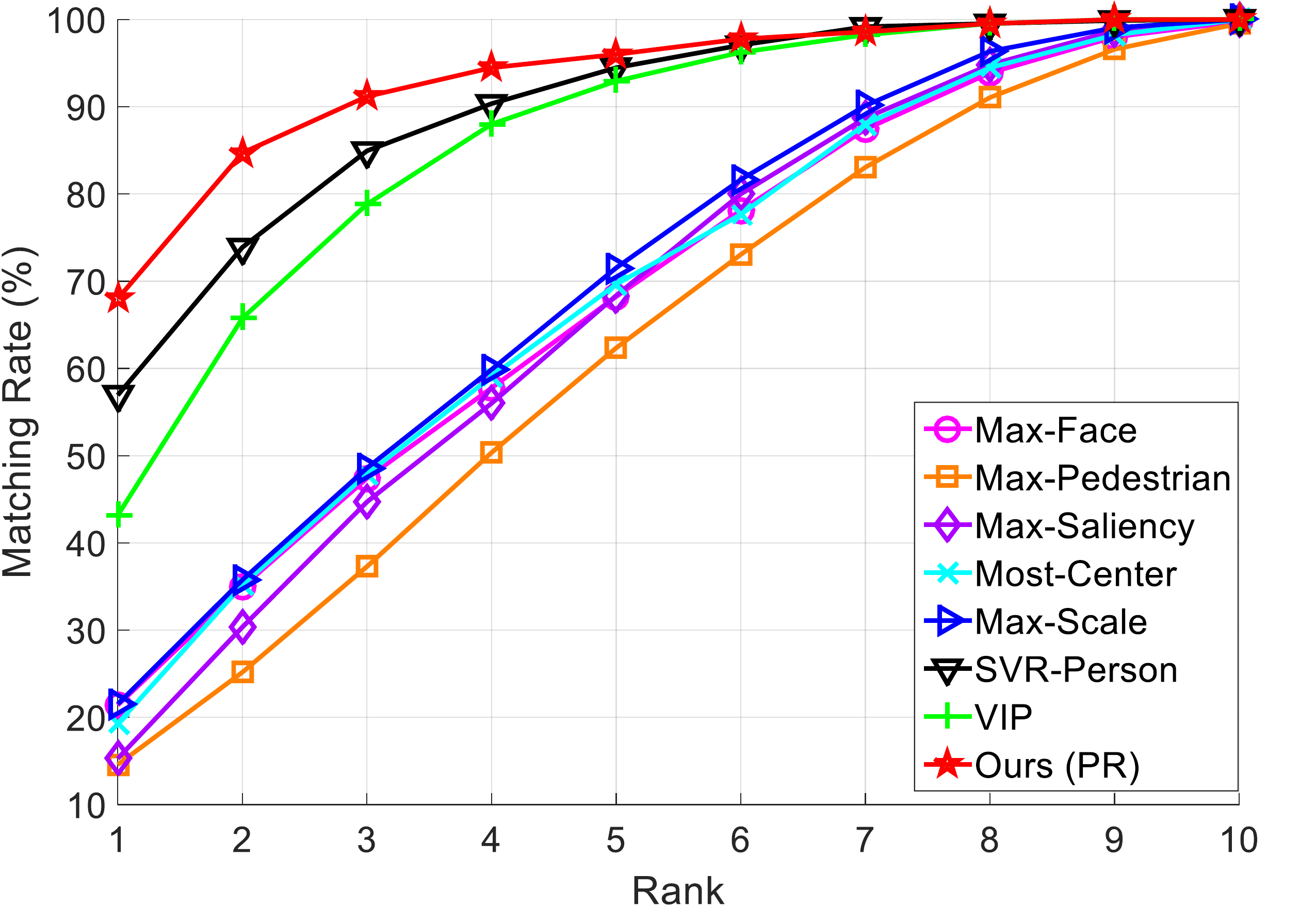}
				}%\vskip -0.3cm
			\vspace{-0.2cm}
				\centering\small\caption{The CMC curve of different methods on two datasets. (Better viewed in color.)}
				\label{fig:cmc}
				\vspace{-0.55cm}
			}
			
		\end{center}
	\end{figure}

	In addition, we report the CMC curve in Figure \ref{fig:CMCcurve}. As shown in Table \ref{tab:Multi-scene}, there are more than 8 persons in an image on average in our Multi-scene Image Dataset. Our approach obtained \whll{85.6\%} \whnew{rank-1} matching rate and \whll{94.2\%} \whnew{rank-2} matching rate. This shows that our proposed method is capable of clearly better inferring \WH{the} importance of persons and localizing the most important one in an image.
	
%	In addition, we also report the CMC curve in Figure \ref{fig:CMCcurve}. As shown in Table \ref{tab:Multi-scene}, there are more than 8 persons in an image on average in our Multi-scene Image Dataset. Our approach obtained 84.1\% \whnew{rank-1} matching rate and 94.5\% \whnew{rank-2} matching rate. This shows that our proposed method is capable of clearly better inferring importance of persons and localizing the most important one in an image.
	\begin{comment}
	\begin{figure}[t]
	\begin{center}
	\label{fig:CMCcurve}
	%\fbox{\rule{0pt}{2in}\rule{0.9\linewidth}{0pt}}
	\includegraphics[width=0.7\linewidth]{figure/CMCFigure.pdf}
	\centering\small\caption{The CMC curve of different methods on Multi-scene important people dataset. (Better viewed in color.)}
	\label{fig:CMCcurve}
	\end{center}
	\end{figure}
	\end{comment}
	
	\vspace{-0.15cm}
	\subsection{Experiments on NCAA Basketball Image Dataset}
	\vspace{-0.2cm}
	
%	\whll{
	The NCAA Basketball Image dataset has its special characteristic that it was captured at a distance so that images are low-resolution and people in images are small.
	\WH{Therefore, the person body bounding box is used to locate a person. In addition, the detail of attention feature and some facial information which  are hard to capture due to the small size of persons are not estimated. So the attention message-based graph will not be used and the spatial graph will be built on feature of person body bounding.}

	\begin{table}[tp]
		\centering
		\footnotesize
		\caption{Mean Average Precison (\%) for Evaluation of Different Methods on NCAA Basketball Image Dataset.}
		\vspace{-0.2cm}
		\resizebox{!}{1.7cm}
		{
		\scriptsize
		\begin{tabular}{c|c|c|c|c|c|c|c|c|c|c}
			
			\hline
			\multirow{2}[2]{*}{Events} & Max-  & Max-  & Max-  & Most- & Max-  & SVR-  & VIP & Ramanathan's & \multirow{1}[2]{*}{\textbf{Ours}} & \multirow{1}[2]{*}{\textbf{Ours}}\\
			& Face  & Pedestrian & Saliency & Center & Scale & Person &  \cite{VIP_solomon2015vip}     & model \cite{Key_ramanathan2015detecting}& \multirow{1}[2]{*}{($\textbf{PR}^{\text{pig}}$)} & \multirow{1}[2]{*}{($\textbf{PR}$)}\\
			\hline
			\hline
			3-point succ. & 35.7 & 29.3 & 12.8 & 14.6 & 26.7 & \color{blue}{56.5} & 47.9 & 51.9 & {\seccolor{67.5}} & {\bf \color{red}{71.0}}\\
			3-point fail. & 32.5 & 27.4 & 15.9 & 12.8 & 24.8 & \color{blue}{58.4} & 48.1 & 54.5 & {\seccolor{71.6}} & {\bf \color{red}{75.2}}\\
			\hline
			free-throw succ. & 33.3 & 37.3 & 13.8 & 11.4 & 63.6 & \color{blue}{86.8} & 55.3 & 77.2 & {\seccolor{89.3}} & {\bf \color{red}{94.4}}\\
			free-throw fail. & 30.9 & 24.1 & 10.1 & 9.6 & \color{blue}{81.8} & 71.7 & 63.9 & 68.5 & {\seccolor{83.9}} & {\bf \color{red}{94.6}}\\
			\hline
			layup succ. & 38.6 & 22.0 & 35.8 & 53.4 & 34.9 & \color{blue}{67.1} & 55.0 & 62.7& {\seccolor{71.1}} & {\bf \color{red}{75.3}}\\
			layup fail. & 32.5 & 23.1 & 37.0 & 44.3 & 41.4 & \color{blue}{64.3} & 55.6 & 60.5& {\seccolor{72.7}} & {\bf \color{red}{74.3}}\\
			\hline
			2-point succ. & 25.6 & 22.1 & 29.9 & 32.2 & 30.7  & \color{blue}{65.9} & 58.6 & 55.4 & {\seccolor{68.1}} & {\bf \color{red}{71.6}}\\
			2-point fail. & 24.8 & 21.2 & 29.8 & 31.3 & 24.8  & \color{blue}{65.9} & 51.6 & 54.2 & {\seccolor{66.6}} & {\bf \color{red}{68.4}}\\
			\hline
			slam dunk succ. & 41.8 & 26.6 & 45.2 & 52.2 & 37.0 & \color{blue}{78.4} & 78.3 & 68.6 & {\bf \color{red}{92.8}} & {\seccolor{89.7}}\\
			slam dunk fail. & 38.5 & 36.5 & 59.4 & \color{blue}{81.3} & 40.6 & {\bf \color{red}{100.0}} & 59.4 & 64.5 & \seccolor{81.3} & \seccolor{81.3}\\
			\hline
			\hline
			\whll{Total}   & 31.4 & 24.7 & 26.4 & 30.0 & 31.8 & \color{blue}{64.5} & 53.2 & 61.8 & {\seccolor{71.1}} & {\bf \color{red}{74.1}}\\
%			mAP   & 31.4 & 24.7 & 26.4 & 30.0 & 31.8 & \color{blue}{64.5} & 53.2 & 61.8 & {\bf \color{magenta}{71.1}} & {\bf \color{red}{74.1}}\\
			\hline
		\end{tabular}%
	}
		\vspace{-0.65cm}
		\label{tab:NCAA_methods}%
	\end{table}%
	
	%\noindent\textbf{Compared to Related Methods}
	
	We report the comparison on CMC curve in Figure \ref{fig:NCAACMC}. In particular, our approach achieved \whll{68.0\%} rank-1 matching rate and \whll{84.7\%} rank-2 matching rate, which indicates that our method can clearly better identify the important people \whll{(shooter)} in the basketball games, whereas the rank-2 matching rate of ``Max-Saliency'', ``Most-Center'', ``Max-Scale'', ``SVR-Person" and VIP is 30.4\%, 35.2\%, 35.8 \%, 73.9\% and 65.8\%, respectively.
	
%	We report the comparison on CMC curve in Figure \ref{fig:NCAACMC}. In particular, our approach achieved 64.0\% rank-1 matching rate and 82.0\% \whnew{rank-2} matching rate, which indicates that our method can clearly better identify the important people in the basketball games, whereas the \whnew{rank-2} matching rate of ``Max-Saliency'', ``Most-Center'', ``Max-Scale'', ``SVR-Person" and VIP is 30.4\%, 35.2\%, 35.8 \%, 73.9\% and 65.8\%, respectively.

	\whll{In addition, we compared a state-of-the-art method reported by Ramanathan et al. \cite{Key_ramanathan2015detecting}, and we call it the Ramanathan's model. 
	Ramanathan's model is not compared in our Multi-scene Important People Image dataset because it can be only run on video-based data \WH{and trained with event labels rather than importance annotations. }
	Although NCAA is a still image-based dataset, it is a subsampling of the original video-based dataset, namely multi-person event detection video dataset  \cite{Key_ramanathan2015detecting}. 
	Hence we compared our method learned on still images with the Ramanathan's model leaned on the full video sequences. }
	%Since no CMC curve result was reported in \cite{Key_ramanathan2015detecting} and instead Ramanathan et al. \cite{Key_ramanathan2015detecting} used mAP which computes the mean of average precision of each event, different from the \whnew{rank-1} matching rate in CMC, we also report the mAP comaprison in Table \ref{tab:NCAA_methods}. 
	Ramanathan's model \wwhl{using} temporal information for important people detection obtained 61.8 \% mAP,
	%mean average precision, 
	and in comparison, our approach achieved 74.1\%. 
	In particular, our approach \wsss{is a graph-based model
%		/***Jason: is feifei's also a graph?***/ 
		and further analyzes hyper-interaction between persons rather than just the pairwise ones}, and in addition exploits some features (such as action features) which are significant but not estimated in Ramanathan's model.
	%, in which only spatial and appearance information are explored in \cite{Key_ramanathan2015detecting}. 
	Also, from the mAP comparison aspect, our proposed method outperformed notably the baseline methods, VIP and ``Max-Saliency'' model again. 
	Our approach outperformed these methods in almost every category and especially performed very well on picking the shooter which is the most important person in the cases like ``free-throw succ./fail"
%	, ``slam dunk succ./fail" 
	and ``3-point succ./fail".

		\begin{figure*}[!htp]
		\begin{center}
		\label{fig:exampleImage}
%		\fbox{\rule{0pt}{2in}\rule{0.9\linewidth}{0pt}}
		\includegraphics[width=0.8\linewidth]{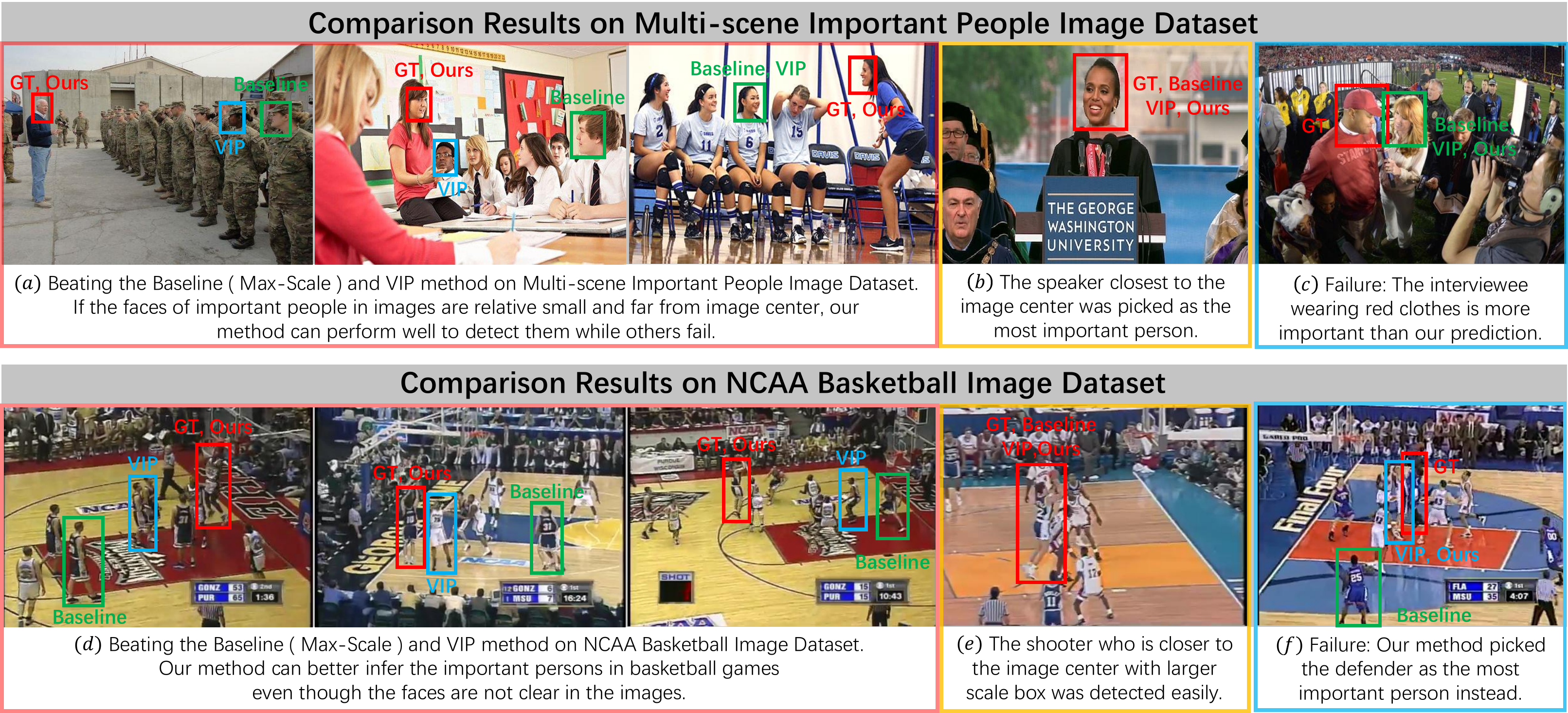}
		\vspace{-0.2cm}
		\centering\small\caption{Some Comparison Results on Multi-scene Important People Image Dataset and NCAA Basketball Image Dataset. (Better viewed in color)}
		\vspace{-0.8cm}
		\label{fig:exampleImage}
		\end{center}
		\end{figure*}
	
	\vspace{-0.15cm}
	\subsection{Visual Comparison}
	\vspace{-0.2cm}
	Figure \ref{fig:exampleImage} shows some qualitative visual results of the baseline (``Max-Scale''), VIP and our approach on the two datasets.
	We can see that VIP and ``Max-Scale'' often picked the person whose bounding box is closest to the image center or the largest one as the most important person. 
	In comparison, as some results shown in Figure \ref{fig:exampleImage} (a) and (d), our approach can detect the important people in images more precisely, although the bounding box of the most important person in an image is not closest to the image center or the largest one. 
	In addition, we show some failure cases (Figure \ref{fig:exampleImage} (c) and (f)). In Figure \ref{fig:exampleImage} (c), the interviewee in the image is the most important person while our method picked the reporter who is interviewing as the most important person.
	In Figure \ref{fig:exampleImage} (f), our method detected the defender that is closest to the shooter as the most important person, while the shooter is the groundtruth. 
	Interestingly, we examined the results and found that the ground-truth most important person annotated was the second most important person output by our model in each of the two images.

%	Figure \ref{fig:exampleImage} shows some qualitative visual results of the baseline (``Max-Scale''), VIP and our approach on the two datasets.
%	We can see that VIP and ``Max-Scale'' often picked the person whose bounding box is \whnew{closest to the image center or the largest one} as the most important person. In comparison, as some results shown in Figure \ref{fig:exampleImage} (a) and (d), our approach can detect the important people in images more precisely, although the bounding box of the most important person in an image is not \whnew{closest} to the image center or the largest one. In addition, we show some failure cases (Figure \ref{fig:exampleImage} (c) and (f)). In Figure \ref{fig:exampleImage} (c), \weihong{the interviewee in the image is the most important person while our method picked the reporter who is interviewing as the most important person.} 
%	In Figure \ref{fig:exampleImage} (f), our method detected the defender that is \whnew{closest} to the shooter as the most important person, while the shooter is the groundtruth. 
%	Interestingly, we examined the results and found that the ground-truth most important person annotated was the second most important person output by our model in each of the two images.

%	\vspace{-0.1cm}

\vspace{-0.1cm}
\begin{table}[tp]
	\centering
	\footnotesize
	\caption{\footnotesize Mean Average Precison (\%) for Evaluation of PR on Both Datasets}
	\vspace{-0.2cm}
	\resizebox{!}{1cm}
	{
		\scriptsize
		\begin{tabular}{c||c|c|c|c|c|c|c|c||c|c|c|c|c|c}
			\hline
			Dataset & \multicolumn{8}{|c||}{Multi-scene Important People Image Dataset} & \multicolumn{6}{c}{NCAA Basketball Image Dataset} \\
			\hline
			Scene & Lecture/ & Demons- & Inter- & \multirow{2}[2]{*}{Sports} & Mili- & Meet- & \multirow{2}[2]{*}{Other} & \multirow{2}[2]{*}{Total} & \multirow{2}[2]{*}{3-point} & free- & \multirow{2}[2]{*}{layup} & \multirow{2}[2]{*}{2-point} & slam & \multirow{2}[2]{*}{Total} \\
			/Events & Speech & stration & view  &       & tary  & ing   &       &       &  & throw &  &  &dunk &  \\
			\hline
			$\text{PR}^{s}$  & 77.8  & 86.0  & 84.7  & 76.4  & 69.5  & 66.1  & 75.8  & 78.7  & 51.7  & 38.1  & 59.2  & 39.2  & 78.6  & 40.5  \\
			\hline
			$\text{PR}^{ap}$ & 79.0  & 80.7  & 79.5  & 76.7  & 78.5  & 52.2  & 77.0  & 77.9  & 65.4  & 90.9  & 68.0  & 62.6  & 82.4  & 59.5  \\
			\hline
			$\text{PR}^{ac}$ & 79.0  & 84.0  & 83.4  & 75.2  & 74.5  & 60.0  & 75.4  & 78.7  & 67.5  & 82.1  & 67.8  & 63.9  & 85.3  & 60.1  \\
			\hline
			$\text{PR}^{at}$ & 83.6  & 84.0  & 86.6  & 68.8  & 80.7  & 77.3  & 83.5  & 82.0  &   -    &  -     &    -   &     -  &    -   & - \\
			\hline
			$\text{PR}$ & {\bf90.2}  & {\bf92.0}  & {\bf90.2}  & {\bf83.6}  & {\bf86.5}  & {\bf76.5}  & {\bf86.7}  & {\bf88.6}  & {\bf73.8}  & {\bf94.5}  & {\bf74.8}  & {\bf69.3}  & {\bf88.5}  & {\bf74.1}  \\
			\hline
		\end{tabular}%
	}
	\vspace{-0.5cm}
	\label{tab:PR_types}%
\end{table}%
	\vspace{-0.15cm}
	\subsection{More Evaluation of the Proposed PR}
	\vspace{-0.15cm}
	\noindent \textbf{Effect of $\mathcal{H}^{hig}$}.
	\whlll{
	For evaluating the effect of forming
%	 hyper-interaction graph 
	 $\mathcal{H}^{hig}$, the mean average precision of PR ($\text{PR}^{\text{pig}}$) inferring \WH{the} importance of persons on the pairwise-interaction graph $\mathcal{H}^{pig}$ only and the our full model {PR} inferring on \wwhl{Hybrid-Interaction Graph} which is constituted as the union of $\mathcal{H}^{pig}$ and $\mathcal{H}^{hig}$ are reported in Table \ref{tab:comBaseMulti} and Table \ref{tab:NCAA_methods}.
	It is evident that \whlllll{$\text{PR}$} makes improvement of 1.1\% (88.6\% - 87.5\%) on Multi-scene Important People Image Dataset and 3.0\% (74.1\%-71.1\%) on NCAA Basketball Image Dataset by $\text{PR}^{\text{pig}}$.
	These results indicate that important people detection can be benefited from exploiting hyper interaction between persons.
}

	\vspace{0.05cm}
	\noindent \textbf{Evaluation of different components in PR}. We now report the \whlll{mAP} of PR using each type of feature, where $\text{PR}^{s}$, $\text{PR}^{ap}$, $\text{PR}^{ac}$, $\text{PR}^{at}$ \weihong{are our approaches using only spatial, appearance, action and attention feature, respectively,} and the full one ($\text{PR}$) that fuses all message graphs in Table \ref{tab:PR_types}.
	Compared to the results of other methods listed in Table \ref{tab:comBaseMulti}, on Multi-scene Important People Image Dataset, our approach using just single type of feature achieved comparable performance in general. Especially, our approach using spatial feature ($\text{PR}^{s}$) still outperformed VIP and the best baseline ``Max-Scale''. On NCAA Basketball dataset, by comparing the results in Table \ref{tab:NCAA_methods} with the ones in Table \ref{tab:PR_types}, our approach using single feature outperformed most \WH{of} other methods.
	When fusing all message graphs, the results in Table \ref{tab:PR_types} also show that $\text{PR}$ makes great improvement on the two datasets, and this suggests all the message graphs are effective and the combination of them yields the best.

%	\noindent \textbf{Evaluation of different components in PR}. We now report the \WHnew{rank-1 matching rate} of PR using each type of feature, where $\text{PR}^{s}$, $\text{PR}^{ap}$, $\text{PR}^{ac}$, $\text{PR}^{at}$ \weihong{are our approaches using only spatial, appearance, action and attention feature, respectively,} and the full one ($\text{PR}$) that fuses all message graphs in Table \ref{tab:PR_types} and Table \ref{tab:NCAA_PR_types}.
%	Compared to the results of other methods listed in Table \ref{tab:comBaseMulti}, on Multi-scene Important People Image Dataset, our approach using just single type of feature achieved comparable performance in general. Especially, our approach using spatial feature ($\text{PR}^{s}$) still outperformed VIP and the best baseline ``Max-Scale''. On NCAA Basketball dataset, by comparing the results in Table \ref{tab:NCAA_methods}with the ones in Table \ref{tab:NCAA_PR_types}, our approach using single feature outperformed most other methods.
%	When fusing all message graphs, the results in Table \ref{tab:PR_types} and Table \ref{tab:NCAA_PR_types} also show that $\text{PR}$ makes great improvement on the two datasets, and this suggests all the message graphs are effective and the combination of them yields the best.
	
%	\vspace{0.05cm}
	
	\noindent \textbf{Parameter Evaluation}. 
	\whll{
		We evaluate $\alpha$ from 0.01 to 1 with interval 0.01. The mAP of our method ranged from 88.5 \% to 88.8 \%
		on the Multi-scene Important People Image Dataset, where it is 88.6 \% at $\alpha=0.85$;
		the mAP ranged from 73.8\% to 74.1\%
		on NCAA Basketball Image Dataset, where it is 74.1\% at $\alpha=0.85$.
		The results indicate that the parameter $\alpha$ is not sensitive in our modeling. 
		Since $\alpha$ is not sensitive, we empirically set $\alpha=0.85$ on both datasets.
		% so that it is not sacrifice computational cost too much.
		}
	
%	In order to evaluate the influence of the parameter $\alpha$ in Eq. (\ref{eq:PR}), we ranged $\alpha$ from 0 to 1 on the two datasets. 
%	\whnew{
%		The \whnew{rank-1} matching rate in CMC of our method ranged from 83.8\% to 84.3\% on the Multi-scene Important People Image dataset, where it is 84.1\% at $\alpha = 0.85$; the \whnew{rank-1} matching rate in CMC of our method ranged from 63.2\% to 64.4\%, where it is 64.0\% at $\alpha=0.85$. The results indicate that the parameter $\alpha$ is not sensitive in our modeling.}
	
	\vspace{-0.1cm}
	\section{Conclusion}
	\vspace{-0.15cm}
	%We have formed a PersonRank method to detect important people in still images. 
	
	\wwhl{The main contribution is to first cast the important people detection problem as node ranking problem in a graph. We form two types of graphs, namely the bidirectional pairwise-interaction graph and the unidirectional hyper-interaction graph, and desgin four edge message functions to mimic the interactions between persons. A modified pagerank algorithm is applied to rank the nodes (i.e. persons) in the graph. This all forms the proposed PersonRank. Extensive \whnew{evaluations} reported on two new datasets have shown the clearly better performance of PersonRank.}
	%: one was collected from multiple scenes including more than six \weihong{categories} and one was gathered as an image-based dataset from a video-based dataset.

%	The key characteristic of our model is to form the Hyper-Interaction Graph that models interactions between persons and between local regions and person as message communicated between nodes.
%%	The key characteristic of our model is to model the people in an image as nodes in a graph and model the interaction between persons as message communicated between nodes. 
%	Finally eigen-analysis is applied to rank the persons and the one corresponding to the most active node is selected as \weihong{the most important person}.  We have compared related methods extensively and have shown that the PersonRank approach has gained clear improvement.
%	

	%------------------------------------------------------------------------

	{\small
		\bibliographystyle{ieee}
		\bibliography{PR_v4_arxiv}
	}
	
\end{document}